\PassOptionsToPackage{table,xcdraw}{xcolor}
\documentclass[sigconf]{acmart}
\usepackage{multirow}
\usepackage{multicol}
\usepackage{pifont}
\usepackage{graphicx}

\AtBeginDocument{%
  }

\copyrightyear{2023}
\acmYear{2023}
\setcopyright{acmlicensed}\acmConference[MM '23]{Proceedings of the 31st ACM International Conference on Multimedia}{October 29-November 3, 2023}{Ottawa, ON, Canada}
\acmBooktitle{Proceedings of the 31st ACM International Conference on Multimedia (MM '23), October 29-November 3, 2023, Ottawa, ON, Canada}
\acmPrice{15.00}
\acmDOI{10.1145/3581783.3611777}
\acmISBN{979-8-4007-0108-5/23/10}
\begin{document}

\title[One-stage Low-resolution Text Recognition with High-resolution Knowledge Transfer]{One-stage Low-resolution Text Recognition \\ with High-resolution Knowledge Transfer}


\author{Hang Guo}
\affiliation{%
  \institution{Tsinghua Shenzhen International Graduate School, Tsinghua University}
  \city{Shenzhen}
  \country{China}
}
\email{cshguo@gmail.com}

\author{Tao Dai}
\authornote{Corresponding author: Tao Dai (daitao.edu@gmail.com).\\
This work is supported in part by the National Key Research and Development Program of China, under Grant 2022YFF1202104, National Natural Science Foundation of China, under Grant (6230070671, 62171248), Shenzhen Science and Technology Program (Grant No.RCYX20200714114523079, JCYJ20220818101014030,JCYJ20220818101012025), and the PCNL KEY project (PCL2021A07).}
\affiliation{%
  \institution{College of Computer Science and Software Engineering, Shenzhen University}
  \city{Shenzhen}
  \country{China}}
\email{daitao.edu@gmail.com}

\author{Mingyan Zhu}
\affiliation{%
  \institution{Tsinghua Shenzhen International Graduate School, Tsinghua University}
  \city{Shenzhen}
  \country{China}
}
\email{zmy20@mails.tsinghua.edu.cn}

\author{Guanghao Meng}
\affiliation{%
 \institution{Tsinghua Shenzhen International Graduate School, Tsinghua University
 Peng Cheng Laboratory}
  \city{Shenzhen}
  \country{China}}
\email{mgh19@mails.tsinghua.edu.cn}

\author{Bin Chen}
\affiliation{%
 \institution{Department of Computer Science and Technology, Harbin Institute of Technology
 }
  \city{Shenzhen}
  \country{China}}
\email{chenbin2021@hit.edu.cn}

\author{Zhi Wang}
\affiliation{%
 \institution{Tsinghua Shenzhen International Graduate School, Tsinghua University}
  \city{Shenzhen}
  \country{China}}
\email{wangzhi@sz.tsinghua.edu.cn}

\author{Shu-Tao Xia}
\affiliation{%
 \institution{Tsinghua Shenzhen International Graduate School, Tsinghua University
 Research Center of Artificial Intelligence, Peng Cheng Laboratory}
  \city{Shenzhen}
  \country{China}}
\email{xiast@sz.tsinghua.edu.cn}

\renewcommand{\shortauthors}{Guo et al.}

\begin{abstract}
Recognizing characters from low-resolution (LR) text images poses a significant challenge due to the information deficiency as well as the noise and blur in low-quality images. 
Current solutions for low-resolution text recognition (LTR) typically rely on a two-stage pipeline that involves super-resolution as the first stage followed by the second-stage recognition. 
Although this pipeline is straightforward and intuitive, it has to use an additional super-resolution network, which causes inefficiencies during training and testing. Moreover, the recognition accuracy of the second stage heavily depends on the reconstruction quality of the first stage, causing ineffectiveness.
In this work, we attempt to address these challenges from a novel perspective: adapting the recognizer to low-resolution inputs by transferring the knowledge from the high-resolution. Guided by this idea, we propose an efficient and effective knowledge distillation framework to achieve multi-level knowledge transfer.
Specifically, the visual focus loss is proposed to extract the character position knowledge with resolution gap reduction and character region focus, the semantic contrastive loss is employed to exploit the contextual semantic knowledge with contrastive learning, and the soft logits loss facilitates both local word-level and global sequence-level learning from the soft teacher label.
Extensive experiments show that the proposed one-stage pipeline significantly outperforms super-resolution based two-stage frameworks in terms of effectiveness and efficiency, accompanied by favorable robustness.
Code is available at \href{https://github.com/csguoh/KD-LTR}{https://github.com/csguoh/KD-LTR}.
\end{abstract}

\ccsdesc[500]{Computing methodologies~Object recognition}
\ccsdesc[300]{Computer systems organization~Neural networks}

\keywords{low-resolution, text recognition, knowledge distillation}


\maketitle


\section{Introduction}
Scene text recognition (STR) has become increasingly popular in recent years due to its various applications, such as license plate recognition \cite{liem2018fvi,khare2019novel}, autonomous driving \cite{bulatov2021approach}, and so on. However, current STR methods suffer from significant performance degradation when recognizing low-resolution images \cite{WenjiaWang2020SceneTI,JingyeChen2021SceneTT,jia2021ifr}.

\begin{figure}[]
\centering
\includegraphics[width=0.46\textwidth]{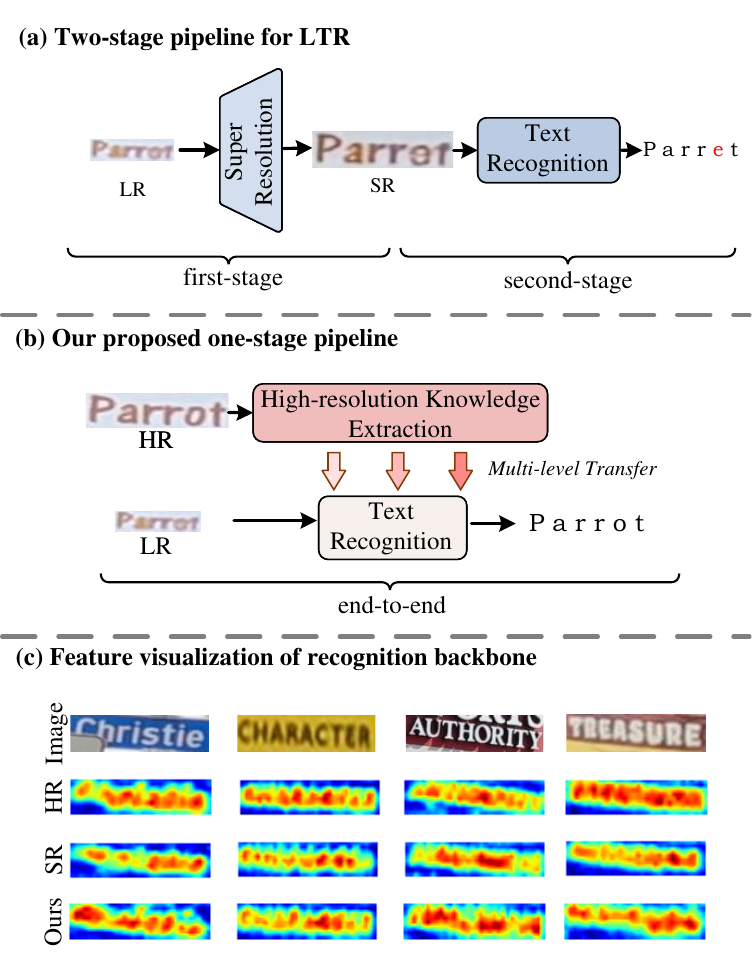}
\caption{(a) The cascading in the two-stage pipeline leads to inefficiency and the error accumulation affects the effectiveness. For example, `Parrot' is incorrectly reconstructed as `Parr{\color[HTML]{CB0000}e}t' in the first stage which then leads to subsequent misrecognition. (b) The proposed framework extracts multi-level knowledge from high-resolution images and transfers it to the recognizer. (c) The features of SR and HR differ in the character regions, e.g. the `i' in `Chr{\color[HTML]{CB0000}i}stie'.}
\label{motivation_fig}
\end{figure}

To address this issue, the mainstream approaches have adopted a two-stage pipeline (see Fig. \ref{motivation_fig} (a)). They split the whole LTR tasks into two separate tasks, generating recognition-friendly text images in the first stage, followed by common text recognition in the second stage. Guided by this two-stage strategy, several pioneering works \cite{dong1506boosting,tran2019deep,xu2017learning} employ the generic single-image super-resolution model to generate SR for recognition. Recently, designing the text-oriented super-resolution model, also known as Scene Text Image Super-Resolution (STISR), has attracted the interest of many researchers \cite{CairongZhao2021SceneTI,JingyeChen2021SceneTT,chen2022text,ma2022text,YongqiangMou2020PlugNetDA,WenjiaWang2020SceneTI,wang2019textsr,nakaune2021skeleton,MinyiZhao2022C3STISRST,qin2022scene,zhu2023improving,ma2023text}. For example, the TextZoom dataset \cite{WenjiaWang2020SceneTI} has been introduced to facilitate real-world STISR research. Furthermore, recent STISR methods \cite{ma2023text,MinyiZhao2022C3STISRST,ma2022text} have utilized pre-trained text recognizers \cite{shi2016end} to inject linguistic knowledge as prior guidance for the super-resolution blocks. Due to the intuitiveness and simplicity, despite some alternative solutions such as multi-task learning \cite{YongqiangMou2020PlugNetDA,jia2021ifr} have been developed, this two-stage pipeline is still prevalent.


As more advanced STISR models continue to be proposed, progress has been made in this two-stage framework. However, this two-stage pipeline also poses certain challenges. First, the two-stage approach necessitates an additional super-resolution network, resulting in inevitably high computational costs for both training and inference. For example, the current state-of-the-art STISR model \cite{MinyiZhao2022C3STISRST} is even larger than the recognition model \cite{bautista2022scene}. Moreover, due to severe information loss and noise in LR, even the use of text-customized super-resolution models may lead to wrong reconstruction and this reconstruction error will further be amplified due to the cascading design. We also visualize the features extracted from SR and HR in the recognizer backbone (see Fig. \ref{motivation_fig} (c)). It can be seen that even though the super-resolution model attempts to imitate HR in the pixel space, there are still differences between both in the feature space.

To break the limitations brought by the two-stage approach, this work explores a one-stage solution by directly adapting the text recognizer to low-resolution inputs without any super-resolution as pre-processing (see Fig. \ref{motivation_fig} (b)). In concrete, we design a LTR-customized knowledge distillation paradigm to mine the knowledge contained in high-resolution images and transfer it to the text recognizer. The key to designing such a paradigm is to find what knowledge should be used and how to transfer this knowledge. To this end, we develop KD-LTR, a novel Knowledge Distillation based framework for LTR, which can help the recognizer learn from high-resolution images. The proposed distillation pipeline extracts three distinct levels of knowledge to facilitate knowledge transfer. Specifically, we employ the visual focus loss to mine the character position knowledge using the resolution gap reduction techniques and mask distillation strategy. In addition, we introduce semantic contrastive loss which uses the contrastive learning scheme to enable the recognizer to acquire contextual semantic knowledge. Finally, the knowledge contained in the soft teacher label is learned from both local word-level and global sequence-level perspectives through the proposed soft logits loss. By leveraging the extensive knowledge encapsulated in high-resolution images from multiple levels, our approach achieves faster and more accurate performance than the super-resolution based two-stage framework and can be easily applied to various text recognition models. 

Overall, our main contributions are three folds:

\begin{itemize}
    \item We propose the first one-stage pipeline for LTR, which adapts the text recognizer to low-resolution inputs by transferring knowledge from high-resolution images.

    \item We propose three well-designed distillation losses to facilitate multi-level knowledge transfer.

    \item Extensive experiments show that the proposed one-stage pipeline sets new state-of-the-art for LTR tasks in terms of efficiency and effectiveness.
\end{itemize}

\begin{figure*}[t]
\centering
\includegraphics[width=0.98\textwidth]{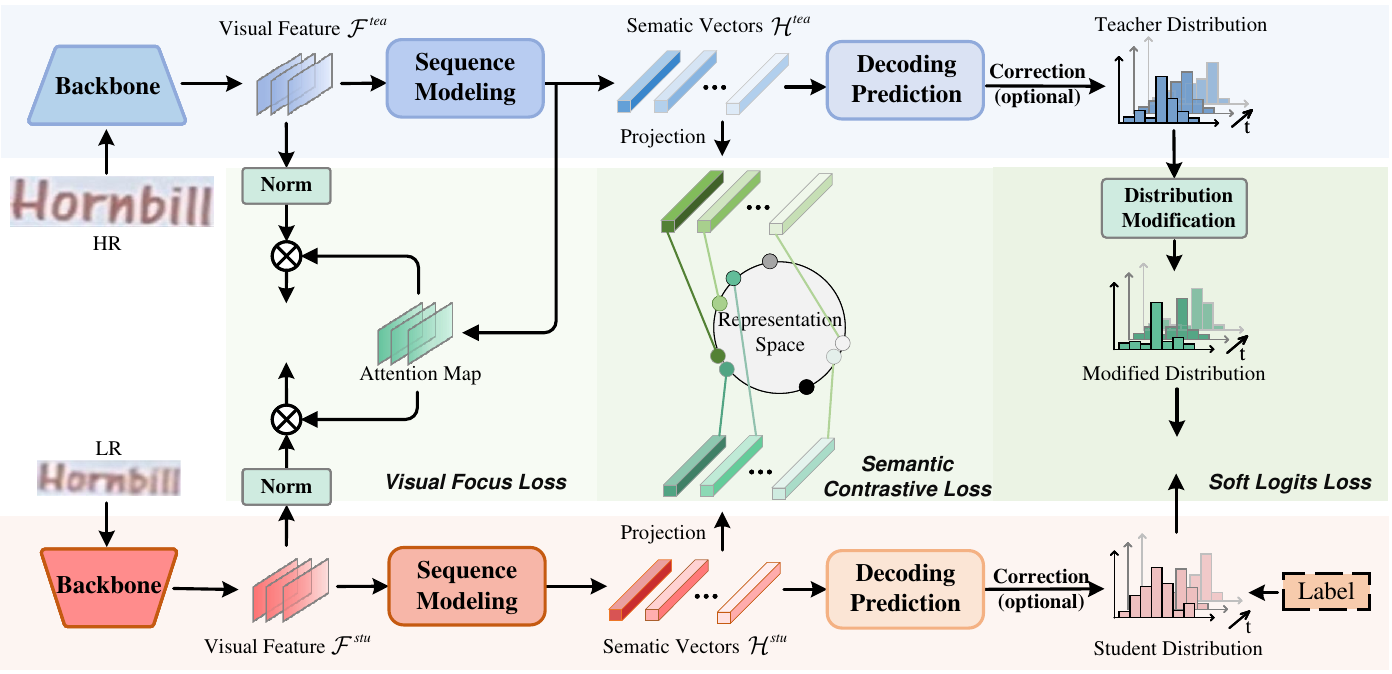}
\caption{An overview of the proposed framework. The branch with {\color[HTML]{759FCC}blue} background is the HR teacher and the branch with {\color[HTML]{FF8C8C}red} background is the LR student. Three levels of loss are used to extract and transfer the knowledge in high-resolution images.}
\label{architecture}
\end{figure*}

\section{Related Work}

Although many vision tasks, including image classification\cite{simonyan2014very,he2016deep}, object detection \cite{carion2020end,redmon2016you}, face recognition \cite{alansari2023ghostfacenets,deng2019arcface}, and text recognition \cite{fang2021read,bautista2022scene} have achieved remarkable success, current methods still struggle with significant performance degradation when confronted with low-resolution images. Depending on the way in which the HR prior is transferred to the LR images, current solutions to this problem can be classified into two types: super-resolution based and knowledge distillation based approaches. Super-resolution based approaches \cite{flusser2015recognition,tran2017disentangled,wang2019textsr,chen2022super} learn the HR prior in the pixel space by employing an additional super-resolution model before visual recognition. In contrast, knowledge distillation based approaches \cite{qi2021multi,ge2020look} transfer knowledge from HR teacher to LR student in the feature space using losses from different perspectives.

\paragraph{Super-resolution for Low-resolution Recognition}
\label{sec:SR-method}
A natural idea to handle the low-resolution visual recognition task is to enhance the original low-quality input to an easily recognizable one by pre-processing. Guided by this idea, some early works borrow from generic Single Image Super-Resolution (SISR) \cite{ledig2017photo,zhang2018image,dai2019second} models to super-resolve the LR to SR before recognition. However, since these super-resolution models are usually trained with image quality as the objective, their effectiveness is limited due to the mismatch in objectives with the recognition task. Recently, there has been a surge of interest in designing task-specific super-resolution models. In the context of scene text recognition, several scene text image super-resolution methods have been proposed, which have shown promising results. For instance, TextSR \cite{wang2019textsr} utilizes a GAN-based architecture to generate recognition-friendly SR images. To facilitate real-world STISR research, the TextZoom dataset \cite{WenjiaWang2020SceneTI} was introduced, accompanied by TSRN which considers the sequential nature of text image data. Moreover, STT \cite{JingyeChen2021SceneTT} implicitly makes the model focus on the character regions by designing relevant loss functions. TATT \cite{ma2022text} achieves spatial deformation robust STISR by using the proposed  TP Interpreter. C3-STISR \cite{MinyiZhao2022C3STISRST} uses three-level clues to guide the super-resolution block and obtains favorable results.

\paragraph{Knowledge Distillation for Low-resolution Recognition}
\label{sec:KD-method}
The concept of Knowledge Distillation (KD) was first introduced by Hinton \textit{et al.} \cite{hinton2015distilling} to transfer knowledge from the over-parameterized teacher to the compact student. More recently, resolution distillation has been developed to address the challenges of low-resolution visual recognition tasks. Unlike traditional KD, which focuses on model compression, resolution distillation transfers knowledge from the HR teacher to the LR student, enabling the student to recover lost information in LR with supervision from different perspectives. While this pipeline has been proven to be effective in image classification \cite{zhu2019low}, object detection \cite{qi2021multi}, and face recognition \cite{shin2022teaching}, it has not been explored for low-resolution text recognition task, whose data involves sequential nature.

\section{Method}

\subsection{Overview}
As shown in Fig. \ref{architecture}, we propose a knowledge distillation framework that can extract different levels of high-resolution knowledge and transfer them to a text recognizer to achieve low-resolution text recognition without additional super-resolution modules. The proposed framework consists of two branches: the HR teacher branch and the LR student branch. The HR teacher branch, which can be obtained from any off-the-shelf text recognizer, takes HR as input and is frozen during training to ensure optimal knowledge transfer. Meanwhile, the LR student branch works with low-resolution text images and aims to recover lost details in the LR input. We begin by reviewing the generic framework for text recognition (Section \ref{sec:text-rec}). Then, we present the details of the loss functions in the proposed distillation framework (Section \ref{sec:adapt-text}).

\subsection{Base Text Recognizer}
\label{sec:text-rec}
The text recognizer used in both the student and teacher branches exploits the prevalent encoder-decoder framework \cite{fang2021read, na2022multi, bautista2022scene}. As shown in each branch of Fig. \ref{architecture}, the text recognition model is broadly divided into three parts: the backbone for feature extraction, the sequence modeling module, and the decoding prediction module. Given input text images $I$, the feature extraction backbone (Resnet \cite{he2016deep} or ViT \cite{dosovitskiy2020image}) first extracts from $I$ to obtain the visual features $\mathcal{F}\in \mathbb{R}^{N\times C \times H \times W}$, where $N$ is the batch size, $C$ is the number of channels, and $H$ and $W$ are the height and width of the features, respectively. Subsequently, the sequence modeling module captures the contextual dependencies with LSTM \cite{hochreiter1997long} or Transformer \cite{vaswani2017attention} to project the 2D visual features $\mathcal{F}$ into 1D semantic vectors $\mathcal{H} \in \mathbb{R}^{N \times T \times C}$, where $T$ is the predefined maximum length of the character sequence. The $\mathcal{H}$ is then transformed into the character probability distribution by the decoding prediction module, which consists of projection layers and a softmax activation function. Linguistic knowledge can be optionally incorporated via an additional language model \cite{fang2021read, na2022multi}.

\subsection{High-resolution Knowledge Transfer}
\label{sec:adapt-text}
To efficiently transfer knowledge from high-resolution images, we propose three levels of loss functions. The first level, visual focus loss, is proposed to extract character position knowledge from the visual features. It can bridge the resolution gap between the two branches and give the character region more focus. The second level, semantic contrastive loss, is exploited to extract contextual semantic knowledge. It facilitates the generation of distinct semantic vectors by leveraging contrastive learning. Finally, the third level, soft logits loss, combines both word-level and sequence-level knowledge from the soft teacher label to produce meaningful recognition results. Further details of these three losses are as follows.

\subsubsection{Visual Focus Loss.}
The visual features extracted from the backbone in the teacher branch contain rich character position knowledge to help the student model recover critical character region features which are useful for subsequent recognition. However, due to the resolution gap of the inputs between the two branches, the visual features of the two are not inherently identical. As such, aligning the student and teacher visual features with absolute numerical measures (e.g., L1 or L2) would be counterproductive. Following \cite{shin2022teaching}, we use the cosine similarity to measure the directional consistency between features. Furthermore, since statistics such as the mean and variance of a feature map can represent the corresponding domain \cite{long2015learning}, we thus first normalize visual features to achieve resolution-domain removal:

\begin{equation}
\tilde{\mathcal{F}} = \frac{\mathcal{F}-\mu}{\sigma},
\end{equation}

\noindent
where $\mathcal{F}$ is a unified notation of the teacher features $\mathcal{F}^{tea}$ and the student features $\mathcal{F}^{stu}$. $\mu$ and $\sigma \in \mathbb{R}^{N \times C}$ are the results of global average pooling and standard deviation pooling of $\mathcal{F}$. 

Then the cosine similarity can be used to measure the distance between the normalized features:

\begin{equation}
\mathcal{L}'_{visual} = 1-\frac{1}{N} \frac{1}{C} \sum_{i=1}^{N} \sum_{j=1}^{C} {\langle \tilde{\mathcal{F}}_{i,j}^{tea},\tilde{\mathcal{F}}_{i,j}^{stu} \rangle},
\end{equation}

\noindent
where $\langle \cdot,\cdot \rangle$ denotes the vector cosine-similarity.

Further, based on the observation in Fig. \ref{motivation_fig} (c), character features are more difficult to learn as well as more important for recognition than background features. Drawing inspiration from \cite{yang2022masked}, we utilize the mask distillation strategy to make the student more focused on the reconstruction of character features while blocking out the disturbance from irrelevant background noise. Specifically, we use the attention map from the teacher branch as a soft mask to reassign the weights of different pixels. Then the visual focus loss with mask distillation can be written as follow:

\begin{equation}
\mathcal{L}_{visual} = 1-\frac{1}{N} \frac{1}{C} \sum_{i=1}^{N} \sum_{j=1}^{C} {\langle \mathcal{M}\tilde{\mathcal{F}}_{i,j}^{tea},\mathcal{M}\tilde{\mathcal{F}}_{i,j}^{stu} \rangle},
\end{equation}

\noindent 
where $\mathcal{M}$ denotes the mask from the teacher attention map. Since most of the popular recognizers \cite{fang2021read,na2022multi,yu2020towards,wang2021two,qiao2021pimnet,xie2022toward} are attention-based, it is easy to obtain the attention map. 

\subsubsection{Semantic Contrastive Loss.}
While the visual focus loss can assist in recovering the lost spatial details in LR, the semantic vectors generated from the sequence modeling module can provide contextual information which is useful for sequential tasks. We therefore exploit this contextual semantic knowledge contained in the HR teacher's semantic vectors $\mathcal{H}^{tea}$.

Specifically, we first obtain the corresponding semantic vector of each character from the recognizer's sequence modeling module:

\begin{equation}
\mathcal{H} = {\rm softmax}(\frac{QK^T}{\sqrt{C}})V,
\end{equation}


\noindent
where $Q$ is the position query of character orders. $K$ and $V$ are the key and value generated from visual features $\mathcal{F}$. The above attentional sequence modeling can make the $i$-th element $h_i$ in $\mathcal{H}$ represent the semantic information corresponding to the $i$-th character in the text image.

We then embed the contrastive learning scheme between $\mathcal{H}^{stu}$ and $\mathcal{H}^{tea}$ to facilitate the learning of more discriminative semantic knowledge. In concrete, we first construct positive and negative samples for contrastive learning. Given $h_i$ in $\mathcal{H}^{stu}(\mathcal{H}^{tea})$ as the anchor, its corresponding positive sample is the $i$-th vector $h'_i$ in the other set $\mathcal{H}^{tea}(\mathcal{H}^{stu})$, and its negative samples are all the elements from the union of $\mathcal{H}^{tea}$ and $\mathcal{H}^{stu}$ after dropping $h_i$ and $h'_i$. Then the contrastive learning is implemented in the 1D semantic representation space:

\begin{equation}
\mathcal{L}_{semantic}=-\frac{1}{L}\sum_{i=1}^{L} \log \frac{\exp({sim}(h_i,h'_i)/\tau)}{\sum_{ h\in \mathcal{H}^{tea} \cup \mathcal{H}^{stu} \textbackslash h_i}\exp({sim}(h_i,h)/\tau)},
\end{equation}

\noindent
where $L$ is the sum of valid character length over batch. $sim(\cdot,\cdot)$ is the distance metric and we use the cosine similarity here. $\tau$ is the distillation temperature. 

Different from previous contrastive learning methods in sequential recognition \cite{aberdam2021sequence,liu2022perceiving} which generate contrastive instances in a predefined and fixed manner (e.g., sliding windows or image patches), we choose to perform contrastive learning on the semantic vectors generated from the sequence modeling module. Since the contrastive instances are already aligned in character order by the attention mechanism, we can obtain each contrastive instance in a data-dependent manner, which is more effective for images with arbitrary text orientations.

\subsubsection{Soft Logits Loss.}
The utilization of the knowledge from the soft teacher label is appealing due to its capability to reflect the similarity between characters \cite{jaderberg2014speeding}. Previous studies \cite{bhunia2021text} have enabled this knowledge transfer by minimizing the KL distance between the output distributions of the student and the teacher at each time step. Nonetheless, this word-level distillation is sub-optimal for sequential tasks as it lacks sequence-level supervision. To this end, we modify the teacher distribution to include both local word-level knowledge and global sequence-level knowledge.

Formally, we refer to the formula in \cite{huang2018knowledge}, and the probability of $k$-th character at time step $t$ in the teacher output $p_t^k$ can be revised as the weighted sum of word-level and sequence-level probabilities: 

\begin{equation}
\label{eq:sequnce-level}
\tilde{p}^k_t = (1-\alpha) p_t^k + \alpha \frac{\sum_{\pi_{t}=k} \prod_{t=1}^T 
 p_t^{\pi_{t}}}{\sum_{\pi} \prod_{t=1}^T p_t^{\pi_t}},
\end{equation}

\noindent
where $\pi$ is all possible decoding paths, $\alpha$ is the hyper-parameter that balances the two distributions. The mathematical derivation can be found in the supplementary material.

However, applying the above equation directly for teacher distribution revision is intractable in practice considering the exponential number of all possible paths. Thus, in practical implementation, we apply some techniques for approximation. Specifically, we select paths with the TopK highest path likelihoods using beam search. Moreover, to ensure the representativity of the beam search results, we only use word-level teacher distribution as supervision when the maximum path likelihood is below a given threshold $r$.

The soft logits loss is then defined as the KL distance between the modified teacher distribution $\tilde{p}$ and the student distribution $q$:

\begin{equation}
\mathcal{L}_{logits}= \frac{1}{T}\sum_{t=1}^T \sum_{k=1}^{|\mathcal{A}|} \tilde{p}^k_t \log\frac{\tilde{p}^k_t}{q^k_t},
\end{equation}

\noindent 
where $|\mathcal{A}|$ is the size of alphabet. Similar to the traditional logits distillation \cite{hinton2015distilling}, we also adopt a high distillation temperature to smooth the distribution.

Although some previous works \cite{kim2016sequence,chen2022dynamic} also employ sequence-related knowledge in the soft teacher label, they treat the whole path likelihood as the atomic element and ignore the fine-grained information pertaining to different characters. For instance, easily-confused characters in one sequence should possess a lower confidence score. In contrast, the proposed sequence-level distribution uses the votes of all decoding paths at each character, which is integrated with word-level distribution to facilitate the acquisition of both global and local knowledge from the soft teacher.

\subsection{Overall Loss}
The overall loss is from the following parts: the task-related cross-entropy loss, the visual focus loss that aids in transferring character position knowledge, the semantic contrastive loss that facilitates contextual semantic knowledge, and the soft logits loss that transfers both word-level and sequence-level soft label knowledge.

\begin{equation}
\mathcal{L} = \lambda_1 \mathcal{L}_{ce} + \lambda_2 \mathcal{L}_{visual}+\lambda_3 \mathcal{L}_{semantic}+\lambda_4 \mathcal{L}_{logits},    
\end{equation}

\noindent
where $\lambda_1$\, $\lambda_2$, $\lambda_3$ and $\lambda_4$ are hyper-parameters.

\section{Experiment}

\subsection{Datasets}

\textbf{TextZoom} \cite{WenjiaWang2020SceneTI} contains 17367 LR-HR scene text image pairs for training and 4373 pairs for testing. The test set is divided into three subsets, with 1619 pairs for the easy subset, 1411 pairs for the medium subset, and 1343 pairs for the hard subset.

\noindent
\textbf{ICDAR2013 (IC13)} \cite{karatzas2013icdar} consists of 1015 images for testing, most of which are regular text images. Some of them are under uneven illumination.

\noindent
\textbf{ICDAR2015 (IC15)} \cite{DimosthenisKaratzas2015ICDAR2C} consists of images taken from scenes and has two versions: 1,811 images $\rm{(IC15_S)}$ and 2,077 images $\rm{(IC15_L)}$. We use $\rm{(IC15_S)}$ for experiments.

\noindent
\textbf{CUTE80} \cite{PalaiahnakoteShivakumara2014ARA} consists of 288 images. Most of them are heavily curved but with high resolution.

\noindent
\textbf{Street View Text (SVT)} \cite{KaiWang2011EndtoendST} has 647 images collected from Google Street View. Some of the images are severely corrupted by noise, blur, and low resolution.

\noindent
\textbf{Street View Text Perspective (SVTP)} \cite{TrungQuyPhan2013RecognizingTW} contains 645 images of which texts are captured in perspective views.


\subsection{Implementation Details}
We use 4 NVIDIA TITAN X GPUs to train our model with batch size 128. We experimentally find that the contrastive learning in $\mathcal{L}_{semantic}$ is not sensitive to batch size. Adam \cite{DiederikPKingma2014AdamAM} is used for optimization. We adopt a learning rate of 5e-5, with a decay factor of 0.1 every 25 epochs. We set the hyper-parameter in total loss $\lambda_1=4$, $\lambda_2=2$, $\lambda_3=0.025$, $\lambda_4=20$. The distillation temperatures in $\mathcal{L}_{semantic}$ and $\mathcal{L}_{logits}$ are $0.1$ and $4$, respectively. In the beam search, we use the paths with the top 6 likelihoods as approximations and set the threshold $r=0.1$. We use the proposed distillation protocol on the currently prevalent SoTA text recognizers, namely ABINet \cite{fang2021read}, MATRN \cite{na2022multi} and PARSeq \cite{bautista2022scene}. Since the input images of the student and teacher branches are of different resolutions, we modified the convolution stride (for CNN backbone) or patch sizes (for ViT backbone) to ensure the consistency of the deep visual features between teacher and student. We use the released pre-trained weights to initialize the student and teacher, and fine-tune the student using the proposed distillation framework. We refer to the student model adapted to low-resolution as ABINet-LTR, MATRN-LTR and PARSeq-LTR, respectively.

\subsection{Evaluation Metrics}
We evaluate the model in terms of efficiency and effectiveness. Specifically, we adopt text recognition accuracy to demonstrate the effectiveness of different methods. We utilize Floating Point Operations (FLOPs) and the number of parameters (Params) to present the efficiency.

\begin{table}[]
\centering
\caption{Ablation on visual focus loss. `cos' denotes cos-similarity loss, L2 loss is used when cos is removed. `norm' denotes the mean-variance normalization. `mask' denotes the mask distillation strategy.}
\label{tab:visuall}
\resizebox{\columnwidth}{!}{%
\begin{tabular}{ccc|cccc}
\hline
\multirow{2}{*}{cos} & \multirow{2}{*}{norm} & \multirow{2}{*}{mask} & \multicolumn{4}{c}{Recognition Accuracy$\uparrow$}                                   \\ \cline{4-7} 
                     &                       &                       & Easy             & Medium           & Hard             & avgAcc           \\ \hline
                     &                       &                       & 86.16\%          & 71.72\%          & 55.17\%          & 71.98\%          \\
\ding{52}                 &                       &                       & 86.04\%          & 72.22\%          & 55.25\%          & 72.12\%          \\
\ding{52}                 & \ding{52}                  &                       & 86.20\%          & 71.89\%          & \textbf{55.72\%} & 72.22\%          \\
\ding{52}                 & \ding{52}                  & \ding{52}                  & \textbf{86.91\%} & \textbf{72.36\%} & 55.10\%          & \textbf{72.45\%} \\ \hline
\end{tabular}%
}
\end{table}

\subsection{Ablation Study}
In this section, we conduct an ablation study to demonstrate the effectiveness of each component. We use the widely adopted ABINet as the base recognizer on the TextZoom dataset.

\begin{table}[]
\centering
\caption{Ablation on semantic contrastive loss. We compare with L2 loss without contrastive learning and SeqCLR with manually predefined contrastive instance division manner.}
\label{sematic loss}
\resizebox{\columnwidth}{!}{%
\begin{tabular}{c|cccc}
\hline
\multirow{2}{*}{semantic loss} & \multicolumn{4}{c}{Recognition Accuracy$\uparrow$} \\ \cline{2-5} 
                           & Easy     & Medium   & Hard     & avgAcc  \\ \hline
L2                   & 86.59\%  & 72.09\%  & 54.95\%  & 72.19\% \\
SeqCLR \cite{aberdam2021sequence}           &85.97\%   & 72.09\%     & \textbf{55.92\%}   &72.26\%     \\
\rowcolor[HTML]{EFEFEF} 
Ours                       & \textbf{86.91\%}  & \textbf{72.36\%}  &  55.10\%  & \textbf{72.45\%}\\ \hline
\end{tabular}%
}
\end{table}

\subsubsection{Ablation on Visual Focus Loss}
The proposed visual focus loss is employed to exploit the rich character position knowledge in the visual features of teacher branch. It contains the cosine similarity with the normalization operator to bridge the resolution gap, and the mask distillation strategy to prompt a character-focused feature learning. We conduct ablation to validate the effectiveness of different components in $\mathcal{L}_{visual}$. Table \ref{tab:visuall} shows the results. The direction-related metric results in an average accuracy improvement of 0.14\% and the normalization operation improves accuracy by 0.1\% through removing the resolution differences. The subsequent mask distillation further improves the average accuracy of 0.23\% by character region focus.

\subsubsection{Ablation on Semantic Contrastive Loss}
We utilize the semantic contrastive loss to extract contextual semantic knowledge with contrastive learning. Since the contrastive instances have been aligned by the attentional sequence modeling module, it is feasible to adaptively decide the number of instances according to the text length in different images. We conduct experiments to verify its effectiveness (see Table \ref{sematic loss}). It can be seen that both contrastive based methods (SeqCLR and ours) outperform the non-contrastive one (L2), suggesting that contrastive learning can extract more discriminative context-aware semantic knowledge. Furthermore, the proposed method is more robust to spatially deformed text images benefiting from the adaptive instance division, leading to an average accuracy of 0.19\% higher than SeqCLR which uses a fixed division manner.

\begin{table}[]
\centering
\caption{Ablation on soft logits loss. WKD denotes word-level distillation. SKD denotes the sequence-level distillation.}
\label{logits loss}
\resizebox{\columnwidth}{!}{%
\begin{tabular}{c|cccc}
\hline
                            & \multicolumn{4}{c}{Recognition Accuracy$\uparrow$}                                  \\ \cline{2-5} 
\multirow{-2}{*}{logits loss} & Easy             & Medium           & Hard             & avgAcc           \\ \hline
WKD \cite{bhunia2021text}                     & 85.86\%          & 72.22\%          & 55.40\%          & 72.10\%          \\
SKD \cite{chen2022dynamic}                     & 85.92\%          & 71.58\%          & 54.43\%          & 71.62\%          \\
\rowcolor[HTML]{EFEFEF} 
Ours                        & \textbf{86.91\%} & \textbf{72.36\%} & \textbf{55.10\%} & \textbf{72.45\%} \\ \hline
\end{tabular}%
}
\end{table}

\begin{table}[]
\centering
\caption{Joint effect ablation of the proposed loss functions.}
\label{tab:comination}
\resizebox{\columnwidth}{0.15\columnwidth}{%
\begin{tabular}{ccc|cccc}
\hline
\multirow{2}{*}{visual} & \multirow{2}{*}{semantic} & \multirow{2}{*}{logits} & \multicolumn{4}{c}{Recognition Accuracy$\uparrow$}                                                                              \\ \cline{4-7} 
                        &                          &                         & Easy                        & Medium                      & Hard                        & avgAcc                      \\ \hline
   &     &    & 81.35\% & 68.32\% & 51.08\% & 67.85\% \\
                        &                          & \ding{52}                    & 86.35\%                     & 72.01\%                     & 54.28\%                     & 71.87\%                     \\
                        & \ding{52}                     & \ding{52}                    & 85.73\%                     & 72.01\%                     & 55.40\%                     & 71.99\%                     \\
\ding{52}                    &                          & \ding{52}                    & 86.21\%                     & 72.16\%                     & \textbf{55.42\%}                     & 72.22\%                     \\
\ding{52}                    & \ding{52}                     & \ding{52}                    & \textbf{86.91\%}            & \textbf{72.36\%}            & 55.10\%            & \textbf{72.45\%}            \\ \hline
\end{tabular}%
}
\end{table}

\begin{table*}[]
\centering
\caption{Comparison with state-of-the-art two-stage methods on the TextZoom dataset regarding efficiency and effectiveness. SR refers to using the STISR model for super-resolution before recognition. KD refers to adapting the text recognizer to low resolution with knowledge distillation.}
\label{compare}
\resizebox{\textwidth}{!}{%
\begin{tabular}{c|cc|cc|cccc}
\hline
                                  &                              &                            &                                                                                                 &                                                                                                 & \multicolumn{4}{c}{Recognition Accuracy$\uparrow$}                                                                                                                                  \\
\multirow{-2}{*}{Text Recognizer} & \multirow{-2}{*}{Method}     & \multirow{-2}{*}{Type}     & \multirow{-2}{*}{\begin{tabular}[c]{@{}c@{}}FLOPs$\downarrow$ \\ ($\times 10^9$)\end{tabular}} & \multirow{-2}{*}{\begin{tabular}[c]{@{}c@{}}Params$\downarrow$ \\ ($\times 10^6$)\end{tabular}} & Easy                                     & Medium                                   & Hard                                     & avgAcc                                   \\ \hline
                                  & Bicubic                      & -                          & -                                                                                               & -                                                                                               & 77.52\%                                  & 56.98\%                                  & 42.81\%                                  & 60.23\%                                  \\
                                  & TSRN \cite{WenjiaWang2020SceneTI}                        & SR                         & 6.38                                                                                            & 39.42                                                                                           & 76.10\%                                  & 61.16\%                                  & 45.57\%                                  & 61.90\%                                  \\
                                  & STT \cite{JingyeChen2021SceneTT}                         & SR                         & 6.66                                                                                            & 39.95                                                                                           & 79.80\%                                  & 64.99\%                                  & 48.47\%                                  & 65.40\%                                  \\
                                  & TATT \cite{ma2022text}                       & SR                         & 7.45                                                                                            & 52.68                                                                                           & 80.67\%                                  & 65.77\%                                  & 50.26\%                                  & 66.52\%                                  \\
                                  & C3-STISR \cite{MinyiZhao2022C3STISRST}                    & SR                         & 8.67                                                                                            & 76.08                                                                                           & 81.35\%                                  & 66.90\%                                  & 49.89\%                                  & 67.03\%                                  \\
\multirow{-6}{*}{ABINet \cite{fang2021read}}          & \cellcolor[HTML]{EFEFEF}Ours & \cellcolor[HTML]{EFEFEF}KD & \cellcolor[HTML]{EFEFEF}\textbf{5.46}                                                           & \cellcolor[HTML]{EFEFEF}\textbf{36.74}                                                          & \cellcolor[HTML]{EFEFEF}\textbf{86.91\%} & \cellcolor[HTML]{EFEFEF}\textbf{72.36\%} & \cellcolor[HTML]{EFEFEF}\textbf{55.10\%} & \cellcolor[HTML]{EFEFEF}\textbf{72.45\%} \\ \hline
                                  & Bicubic                      & -                          & -                                                                                               & -                                                                                               & 80.42\%                                  & 58.97\%                                  & 44.97\%                                  & 62.61\%                                  \\
                                  & TSRN \cite{WenjiaWang2020SceneTI}                        & SR                         & 10.74                                                                                           & 46.84                                                                                           & 77.08\%                                  & 62.65\%                                  & 47.21\%                                  & 63.25\%                                  \\
                                  & STT \cite{JingyeChen2021SceneTT}                         & SR                         & 11.06                                                                                           & 47.37                                                                                           & 81.66\%                                  & 65.98\%                                  & 50.11\%                                  & 66.91\%                                  \\
                                  & TATT \cite{ma2022text}                       & SR                         & 11.85                                                                                           & 60.10                                                                                           & 81.10\%                                  & 66.62\%                                  & 51.68\%                                  & 67.39\%                                  \\
                                  & C3-STISR \cite{MinyiZhao2022C3STISRST}                    & SR                         & 13.07                                                                                           & 83.50                                                                                           & 81.90\%                                  & 68.04\%                                  & 51.08\%                                  & 67.96\%                                  \\
\multirow{-6}{*}{MATRN \cite{na2022multi}}           & \cellcolor[HTML]{EFEFEF}Ours & \cellcolor[HTML]{EFEFEF}KD & \cellcolor[HTML]{EFEFEF}\textbf{9.86}                                                           & \cellcolor[HTML]{EFEFEF}\textbf{44.16}                                                          & \cellcolor[HTML]{EFEFEF}\textbf{86.91\%} & \cellcolor[HTML]{EFEFEF}\textbf{73.14\%} & \cellcolor[HTML]{EFEFEF}\textbf{56.96\%} & \cellcolor[HTML]{EFEFEF}\textbf{73.27\%} \\ \hline
                                  & Bicubic                      & -                          & -                                                                                               & -                                                                                               & 90.36\%                                  & 75.34\%                                  & 57.11\%                                  & 75.30\%                                  \\
                                  & TSRN \cite{WenjiaWang2020SceneTI}                         & SR                         & 3.76                                                                                            & 26.51                                                                                           & 79.74\%                                  & 61.09\%                                  & 47.65\%                                  & 63.87\%                                  \\
                                  & STT \cite{JingyeChen2021SceneTT}                        & SR                         & 4.40                                                                                            & 27.04                                                                                           & 83.69\%                                  & 66.69\%                                  & 51.75\%                                  & 68.40\%                                  \\
                                  & TATT \cite{ma2022text}                        & SR                         & 4.83                                                                                            & 39.77                                                                                           & 82.21\%                                  & 65.91\%                                  & 52.12\%                                  & 67.71\%                                  \\
                                  & C3-STISR \cite{MinyiZhao2022C3STISRST}                     & SR                         & 6.06                                                                                            & 63.17                                                                                           & 84.25\%                                  & 68.25\%                                  & 50.86\%                                  & 68.83\%                                  \\
\multirow{-6}{*}{PARSeq \cite{bautista2022scene}}          & \cellcolor[HTML]{EFEFEF}Ours & \cellcolor[HTML]{EFEFEF}KD & \cellcolor[HTML]{EFEFEF}\textbf{2.93}                                                           & \cellcolor[HTML]{EFEFEF}\textbf{23.81}                                                          & \cellcolor[HTML]{EFEFEF}\textbf{90.36\%} & \cellcolor[HTML]{EFEFEF}\textbf{78.88\%} & \cellcolor[HTML]{EFEFEF}\textbf{63.22\%} & \cellcolor[HTML]{EFEFEF}\textbf{78.23\%} \\ \hline
\end{tabular}%
}
\end{table*}

\begin{figure}[t]
\centering
\includegraphics[width=0.99\columnwidth]{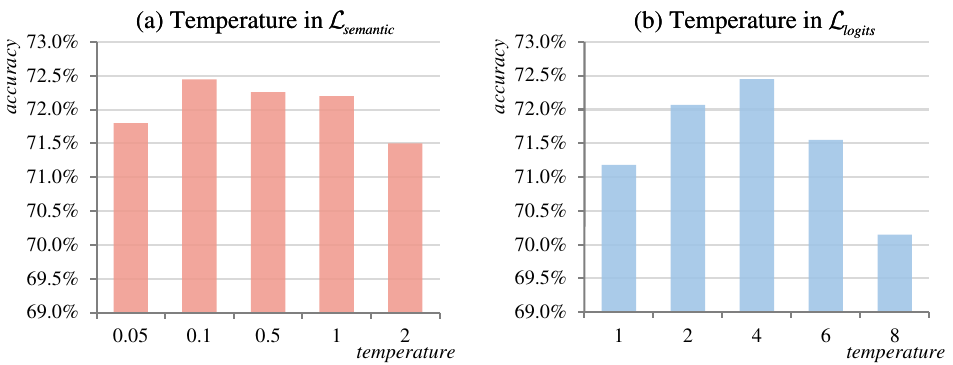}
\caption{Ablation of distillation temperature on (a) semantic contrastive loss and (b) soft logits loss.}
\label{fig:logits-param}
\end{figure}



\subsubsection{Ablation on Soft Logits Loss}
The proposed soft logits loss introduces both local word-level and global sequence-level knowledge from the soft teacher label. To verify the validity, we replace the proposed loss with those used in previous work, i.e. word-level KL loss \cite{bhunia2021text} with each time step modeled independently and sequence distillation loss \cite{chen2022dynamic} with the entire path likelihood as the target. Table \ref{logits loss} presents the results. It can be seen that only sub-optimal results can be achieved with only one single-level loss. By contrast, by providing both local and global knowledge, the proposed method outperforms others.

\subsubsection{Combination of Different Losses}
We analyze the combined effects of the three proposed distillation losses (refer to Table \ref{tab:comination}). The results show that using only the task-related cross-entropy loss yielded an accuracy of 67.85\%. By incorporating the knowledge from the soft teacher label, there is a recognition accuracy improvement of 4.02\%. The addition of semantic contrastive loss, which can migrate contextual semantic knowledge, led to an increase in accuracy of 0.12\%. The visual focus loss, which transfers character position knowledge, further improved the accuracy by 0.35\%. Finally, the best result of 72.45\% was achieved when all three losses were utilized.

\begin{figure}[t]
\centering
\includegraphics[width=0.95\columnwidth]{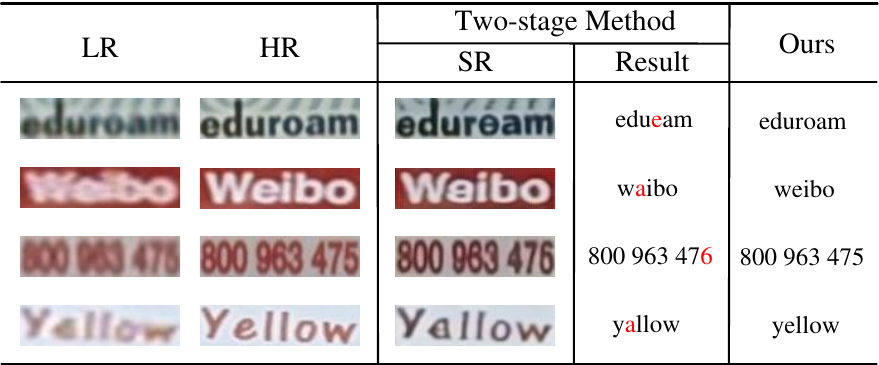}
\caption{Qualitative comparison with the two-stage approach \cite{MinyiZhao2022C3STISRST}. Erroneous reconstruction due to low quality of LR in the two-stage pipeline can affect subsequent text recognition.}
\label{quantitive-conpare}
\end{figure}

\subsubsection{Ablation on Hyper-parameters}
Both semantic contrastive loss and soft logits loss contain temperature hyper-parameters, here we focus on how distillation temperature affects model performance. Fig. \ref{fig:logits-param} (a) shows the effects of distillation temperature in $\mathcal{L}_{semantic}$. It can be seen that as the distillation loss gets smaller, the recognition accuracy increases. This is due to the fact that a small temperature amplifies the differences between the semantic vectors in contrastive learning, forcing the model to focus on the subtle differences, thus facilitating the learning of more discriminative embeddings and leading to an increase in performance. However, too low a temperature can lead to an unhealthy gradient by over-amplifying the difference. In addition, the variation of model performance with temperature in $\mathcal{L}_{logits}$ is shown in Fig. \ref{fig:logits-param} (b). A larger distillation temperature can make the distribution more uniform, thus aiding the learning of similarities between characters and enhancing the model's ability to differentiate between confusable characters. However, too large a temperature is detrimental due to the reduction of information.

\subsection{Comparison to State-of-the-Arts}

We first compare our method with super-resolution based two-stage methods. After that, we conduct the robustness test on five challenging STR benchmarks. Finally, we compare with other solutions to LTR.

\subsubsection{Comparison with Two-stage Methods}
We compare with the super-resolution based two-stage approaches which use the STISR model before recognition. Table \ref{compare} shows the results on the TextZoom dataset. Surprisingly, the recognition accuracy of bicubic even exceeds that of the SoTA
STISR method \cite{MinyiZhao2022C3STISRST} when applying a recognizer \cite{bautista2022scene} with strong
contextual modeling capabilities. We speculate that it is due to the manipulation of the original LR image by the super-resolution model. Moreover, by including only one text recognizer without any pre-processing model, the proposed framework is highly efficient, for example, PARSeq-LTR saves 3.13 GFLOPs and 39.36M parameters compared with C3-STISR. In addition, our method achieves a significant recognition accuracy improvement over the two-stage pipelines due to the use of joint optimization. Impressively, PARSeq-LTR even outperforms the two-stage pipeline which uses C3-STISR model for STISR by 9.4\% on average recognition accuracy. A qualitative comparison is shown in Fig. \ref{quantitive-conpare}.


\subsubsection{Robustness Comparison}
We first test the data distribution shift robustness of different methods by freezing the model trained on TextZoom and directly transferring them to five scene text recognition benchmarks. Since these STR datasets contain high-resolution images, we first manually degrade the raw images, including blur, noise, etc., before using the degraded images to test the robustness, see the supplementary material for details.

Table \ref{STR compare} shows the experimental results. It can be seen that the proposed method achieves SoTA performance on all five STR datasets even after transferring to other data distribution. For example, PARSeq-LTR achieves a 7.51\% performance boost versus C3-STISR on IC15 dataset. The above results demonstrate the data distribution robustness of the proposed method.

\begin{figure}[t]
\centering
\includegraphics[width=\columnwidth]{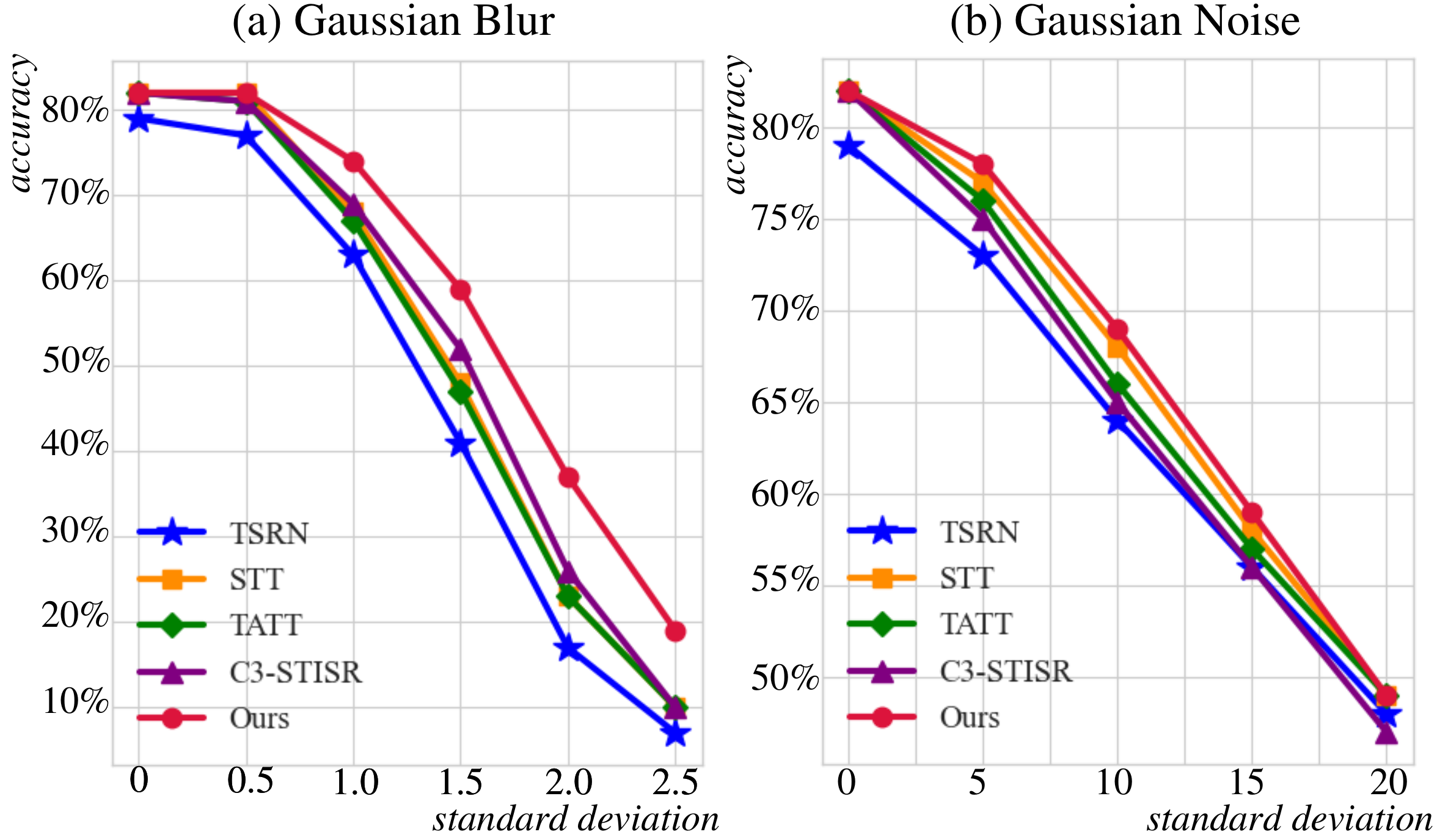}
\caption{Recognition accuracy of different methods with varying (a) Gaussian Blur and (b) Gaussian Noise. ABINet is used on the IC15 dataset.}
\label{blur}
\end{figure}

Furthermore, we study the performance variation of different methods under different levels of Gaussian Blur and Gaussian Noise. Fig. \ref{blur} shows the results. For the gaussian blur, our method drops more slowly and outperforms the super-resolution based two-stage methods at all blur settings. For the gaussian noise, the difference among different methods is not significant when the noise is at a high setting, but our method still beats others.

\begin{table}[]
\centering
\caption{Adaption to scene text recognition benchmarks. Images are manually downscaled to obtain LR for testing.}
\label{STR compare}
\resizebox{\columnwidth}{!}{%
\begin{tabular}{c|ccccc}
\hline
                         & \multicolumn{5}{c}{Recognition Accuracy$\uparrow$}                                                                                    \\ \cline{2-6} 
\multirow{-2}{*}{Method} & IC13             & IC15                            & CUTE80           & SVT              & SVTP                                     \\ \hline
\textit{abinet}          &                  &                                 &                  &                  &                                          \\
TSRN                     & 53.20\%          & 36.44\%                         & 39.93\%          & 42.66\%          & 36.59\%                                  \\
STT                      & 56.35\%          & 39.37\%                         & 41.30\%          & 41.42\%          & 40.47\%                                  \\
TATT                     & 53.60\%          & 36.72\%                         & 40.97\%          & 44.05\%          & 37.36\%                                  \\
C3-STISR                 & 55.17\%          & 39.48\%                         & 36.81\%          & 44.51\%          & 39.22\%                                  \\
\rowcolor[HTML]{EFEFEF} 
Ours                     & \textbf{57.44\%} & \textbf{44.17\%}                & \textbf{41.32\%} & \textbf{46.06\%} & \cellcolor[HTML]{EFEFEF}\textbf{42.64\%} \\ \hline
\textit{matrn}           &                  &                                 &                  &                  &                                          \\
TSRN                     & 54.09\%          & 35.89\%                         & 39.93\%          & 43.43\%          & 38.76\%                                  \\
STT                      & 56.65\%          & \cellcolor[HTML]{FFFFFF}40.36\% & 42.36\%          & 43.89\%          & 42.02\%                                  \\
TATT                     & 57.43\%          & 39.76\%                         & 42.01\%          & 43.12\%          & 43.41\%                                  \\
C3-STISR                 & 57.64\%          & 42.19\%                         & 42.36\%          & 44.51\%          & 42.64\%                                  \\
\rowcolor[HTML]{EFEFEF} 
Ours                     & \textbf{59.01\%} & \textbf{43.57\%}                & \textbf{43.40\%} & \textbf{53.31\%} & \cellcolor[HTML]{EFEFEF}\textbf{43.72\%} \\ \hline
\textit{parsq}           &                  &                                 &                  &                  &                                          \\
TSRN                     & 55.86\%          & 38.98\%                         & 47.92\%          & 44.20\%          & 39.69\%                                  \\
STT                      & 60.89\%          & 42.79\%                         & 46.18\%          & 46.68\%          & 44.03\%                                  \\
TATT                     & 55.96\%          & 41.74\%                         & 48.26\%          & 46.21\%          & 41.24\%                                  \\
C3-STISR                 & 58.72\%          & 42.30\%                         & 44.10\%          & 45.44\%          & 42.33\%                                  \\
\rowcolor[HTML]{EFEFEF} 
Ours                     & \textbf{62.36\%}                & \textbf{49.81\%}                               & \textbf{53.82\%}                & \textbf{53.94\%}                & \textbf{49.61\%}                                       \\ \hline
\end{tabular}%
}
\end{table}

\subsubsection{Comparison with Other Solutions to LTR}
In addition to the super-resolution based two-stage framework, multi-task learning, which handles LTR by learning common visual features for both super-resolution and text recognition, has also been proposed \cite{YongqiangMou2020PlugNetDA,jia2021ifr}. In this work, we compare with two representative multi-task learning based methods, i.e., PlugNet \cite{YongqiangMou2020PlugNetDA} and IFR \cite{jia2021ifr}. To ensure a fair comparison, we apply the plug-and-play feature super-resolution modules of PlugNet and IFR to ABINet, and compare them with our ABINet-LTR. The results presented in Table \ref{table:other-solution} show that although this multi-task framework outperforms the super-resolution based two-stage one, it still falls behind the proposed distillation framework. This is because these methods are limited to pixel space learning as well as do not take into account supervision from different perspectives, such as semantic space and sequence modeling, which results in limited performance.

\begin{table}[]
\centering
\caption{Comparison with multi-task learning based pipelines. The proposed high-resolution knowledge transfer framework still achieves SoTA performance.}
\label{table:other-solution}
\resizebox{\columnwidth}{!}{%
\begin{tabular}{c|cccc}
\hline
                         & \multicolumn{4}{c}{Recognition Accuracy$\uparrow$}                                  \\ \cline{2-5} 
\multirow{-2}{*}{Method} & Easy             & Medium           & Hard             & avgAcc           \\ \hline
PlugNet \cite{YongqiangMou2020PlugNetDA}                  & 81.90\%          & 68.89\%          & 52.27\%          & 68.60\%          \\
IFR \cite{jia2021ifr}         &   82.58\%               &  68.89\%        &   52.87\%    &    69.04\%          \\
\rowcolor[HTML]{EFEFEF} 
Ours                     & \textbf{86.91\%} & \textbf{72.36\%} & \textbf{55.10\%} & \textbf{72.45\%} \\ \hline
\end{tabular}%
}
\end{table}

\section{Conclusion}
We propose a novel knowledge distillation framework that adapts text recognizers for the low-resolution to address challenges posed by previous super-resolution based two-stage pipelines. Three distillation losses are designed to extract multi-level knowledge from the high-resolution. The visual focus loss transfers the character position knowledge in a resolution-agnostic manner and reinforces the character focus with mask distillation. The semantic contrastive loss leverages contrastive learning to facilitate the learning of discriminative contextual semantic knowledge. The soft logits loss models both local word-level and global sequence-level knowledge in the soft teacher label for better supervision. Extensive experiments demonstrate that the proposed one-stage pipeline achieves state-of-the-art performance against two-stage counterparts in terms of efficiency and effectiveness, with favorable robustness. We hope that our work can inspire more studies on one-stage LTR.


\bibliographystyle{ACM-Reference-Format}
\bibliography{acmart}


\begin{thebibliography}{62}


\ifx \showCODEN    \undefined \def \showCODEN     #1{\unskip}     \fi
\ifx \showDOI      \undefined \def \showDOI       #1{#1}\fi
\ifx \showISBNx    \undefined \def \showISBNx     #1{\unskip}     \fi
\ifx \showISBNxiii \undefined \def \showISBNxiii  #1{\unskip}     \fi
\ifx \showISSN     \undefined \def \showISSN      #1{\unskip}     \fi
\ifx \showLCCN     \undefined \def \showLCCN      #1{\unskip}     \fi
\ifx \shownote     \undefined \def \shownote      #1{#1}          \fi
\ifx \showarticletitle \undefined \def \showarticletitle #1{#1}   \fi
\ifx \showURL      \undefined \def \showURL       {\relax}        \fi
\providecommand\bibfield[2]{#2}
\providecommand\bibinfo[2]{#2}
\providecommand\natexlab[1]{#1}
\providecommand\showeprint[2][]{arXiv:#2}

\bibitem[Aberdam et~al\mbox{.}(2021)]%
        {aberdam2021sequence}
\bibfield{author}{\bibinfo{person}{Aviad Aberdam}, \bibinfo{person}{Ron
  Litman}, \bibinfo{person}{Shahar Tsiper}, \bibinfo{person}{Oron Anschel},
  \bibinfo{person}{Ron Slossberg}, \bibinfo{person}{Shai Mazor},
  \bibinfo{person}{R Manmatha}, {and} \bibinfo{person}{Pietro Perona}.}
  \bibinfo{year}{2021}\natexlab{}.
\newblock \showarticletitle{Sequence-to-sequence contrastive learning for text
  recognition}. In \bibinfo{booktitle}{\emph{Proceedings of the IEEE/CVF
  Conference on Computer Vision and Pattern Recognition}}.
  \bibinfo{pages}{15302--15312}.
\newblock


\bibitem[Alansari et~al\mbox{.}(2023)]%
        {alansari2023ghostfacenets}
\bibfield{author}{\bibinfo{person}{Mohamad Alansari},
  \bibinfo{person}{Oussama~Abdul Hay}, \bibinfo{person}{Sajid Javed},
  \bibinfo{person}{Abdulhaid Shoufan}, \bibinfo{person}{Yahya Zweiri}, {and}
  \bibinfo{person}{Naoufel Werghi}.} \bibinfo{year}{2023}\natexlab{}.
\newblock \showarticletitle{GhostFaceNets: Lightweight Face Recognition Model
  from Cheap Operations}.
\newblock \bibinfo{journal}{\emph{IEEE Access}} (\bibinfo{year}{2023}).
\newblock


\bibitem[Bautista and Atienza(2022)]%
        {bautista2022scene}
\bibfield{author}{\bibinfo{person}{Darwin Bautista} {and}
  \bibinfo{person}{Rowel Atienza}.} \bibinfo{year}{2022}\natexlab{}.
\newblock \showarticletitle{Scene text recognition with permuted autoregressive
  sequence models}. In \bibinfo{booktitle}{\emph{Computer Vision--ECCV 2022:
  17th European Conference, Tel Aviv, Israel, October 23--27, 2022,
  Proceedings, Part XXVIII}}. Springer, \bibinfo{pages}{178--196}.
\newblock


\bibitem[Bhunia et~al\mbox{.}(2021)]%
        {bhunia2021text}
\bibfield{author}{\bibinfo{person}{Ayan~Kumar Bhunia},
  \bibinfo{person}{Aneeshan Sain}, \bibinfo{person}{Pinaki~Nath Chowdhury},
  {and} \bibinfo{person}{Yi-Zhe Song}.} \bibinfo{year}{2021}\natexlab{}.
\newblock \showarticletitle{Text is text, no matter what: Unifying text
  recognition using knowledge distillation}. In
  \bibinfo{booktitle}{\emph{Proceedings of the IEEE/CVF International
  Conference on Computer Vision}}. \bibinfo{pages}{983--992}.
\newblock


\bibitem[Bulatov et~al\mbox{.}(2021)]%
        {bulatov2021approach}
\bibfield{author}{\bibinfo{person}{Konstantin Bulatov},
  \bibinfo{person}{Nadezhda Fedotova}, {and} \bibinfo{person}{Vladimir~V
  Arlazarov}.} \bibinfo{year}{2021}\natexlab{}.
\newblock \showarticletitle{An approach to road scene text recognition with
  per-frame accumulation and dynamic stopping decision}. In
  \bibinfo{booktitle}{\emph{Thirteenth International Conference on Machine
  Vision}}, Vol.~\bibinfo{volume}{11605}. SPIE, \bibinfo{pages}{511--519}.
\newblock


\bibitem[Carion et~al\mbox{.}(2020)]%
        {carion2020end}
\bibfield{author}{\bibinfo{person}{Nicolas Carion}, \bibinfo{person}{Francisco
  Massa}, \bibinfo{person}{Gabriel Synnaeve}, \bibinfo{person}{Nicolas
  Usunier}, \bibinfo{person}{Alexander Kirillov}, {and} \bibinfo{person}{Sergey
  Zagoruyko}.} \bibinfo{year}{2020}\natexlab{}.
\newblock \showarticletitle{End-to-end object detection with transformers}. In
  \bibinfo{booktitle}{\emph{Computer Vision--ECCV 2020: 16th European
  Conference, Glasgow, UK, August 23--28, 2020, Proceedings, Part I 16}}.
  Springer, \bibinfo{pages}{213--229}.
\newblock


\bibitem[Chen et~al\mbox{.}(2022a)]%
        {chen2022super}
\bibfield{author}{\bibinfo{person}{Hongyuan Chen}, \bibinfo{person}{Yanting
  Pei}, \bibinfo{person}{Hongwei Zhao}, {and} \bibinfo{person}{Yaping Huang}.}
  \bibinfo{year}{2022}\natexlab{a}.
\newblock \showarticletitle{Super-resolution guided knowledge distillation for
  low-resolution image classification}.
\newblock \bibinfo{journal}{\emph{Pattern Recognition Letters}}
  \bibinfo{volume}{155} (\bibinfo{year}{2022}), \bibinfo{pages}{62--68}.
\newblock


\bibitem[Chen et~al\mbox{.}(2021)]%
        {JingyeChen2021SceneTT}
\bibfield{author}{\bibinfo{person}{Jingye Chen}, \bibinfo{person}{Bin Li},
  {and} \bibinfo{person}{Xiangyang Xue}.} \bibinfo{year}{2021}\natexlab{}.
\newblock \showarticletitle{Scene Text Telescope: Text-Focused Scene Image
  Super-Resolution}.
\newblock \bibinfo{journal}{\emph{computer vision and pattern recognition}}
  (\bibinfo{year}{2021}).
\newblock


\bibitem[Chen et~al\mbox{.}(2022c)]%
        {chen2022text}
\bibfield{author}{\bibinfo{person}{Jingye Chen}, \bibinfo{person}{Haiyang Yu},
  \bibinfo{person}{Jianqi Ma}, \bibinfo{person}{Bin Li}, {and}
  \bibinfo{person}{Xiangyang Xue}.} \bibinfo{year}{2022}\natexlab{c}.
\newblock \showarticletitle{Text gestalt: Stroke-aware scene text image
  super-resolution}. In \bibinfo{booktitle}{\emph{Proceedings of the AAAI
  Conference on Artificial Intelligence}}, Vol.~\bibinfo{volume}{36}.
  \bibinfo{pages}{285--293}.
\newblock


\bibitem[Chen et~al\mbox{.}(2022b)]%
        {chen2022dynamic}
\bibfield{author}{\bibinfo{person}{Ying Chen}, \bibinfo{person}{Liang Qiao},
  \bibinfo{person}{Zhanzhan Cheng}, \bibinfo{person}{Shiliang Pu},
  \bibinfo{person}{Yi Niu}, {and} \bibinfo{person}{Xi Li}.}
  \bibinfo{year}{2022}\natexlab{b}.
\newblock \showarticletitle{Dynamic Low-Resolution Distillation for
  Cost-Efficient End-to-End Text Spotting}. In
  \bibinfo{booktitle}{\emph{Computer Vision--ECCV 2022: 17th European
  Conference, Tel Aviv, Israel, October 23--27, 2022, Proceedings, Part
  XXVIII}}. Springer, \bibinfo{pages}{356--373}.
\newblock


\bibitem[Dai et~al\mbox{.}(2019)]%
        {dai2019second}
\bibfield{author}{\bibinfo{person}{Tao Dai}, \bibinfo{person}{Jianrui Cai},
  \bibinfo{person}{Yongbing Zhang}, \bibinfo{person}{Shu-Tao Xia}, {and}
  \bibinfo{person}{Lei Zhang}.} \bibinfo{year}{2019}\natexlab{}.
\newblock \showarticletitle{Second-order attention network for single image
  super-resolution}. In \bibinfo{booktitle}{\emph{Proceedings of the IEEE/CVF
  conference on computer vision and pattern recognition}}.
  \bibinfo{pages}{11065--11074}.
\newblock


\bibitem[Deng et~al\mbox{.}(2019)]%
        {deng2019arcface}
\bibfield{author}{\bibinfo{person}{Jiankang Deng}, \bibinfo{person}{Jia Guo},
  \bibinfo{person}{Niannan Xue}, {and} \bibinfo{person}{Stefanos Zafeiriou}.}
  \bibinfo{year}{2019}\natexlab{}.
\newblock \showarticletitle{Arcface: Additive angular margin loss for deep face
  recognition}. In \bibinfo{booktitle}{\emph{Proceedings of the IEEE/CVF
  conference on computer vision and pattern recognition}}.
  \bibinfo{pages}{4690--4699}.
\newblock


\bibitem[Dong et~al\mbox{.}(2015)]%
        {dong1506boosting}
\bibfield{author}{\bibinfo{person}{C Dong}, \bibinfo{person}{X Zhu},
  \bibinfo{person}{Y Deng}, \bibinfo{person}{CC Loy}, {and} \bibinfo{person}{Y
  Qia}.} \bibinfo{year}{2015}\natexlab{}.
\newblock \showarticletitle{Boosting optical character recognition: A
  super-resolution approach}.
\newblock \bibinfo{journal}{\emph{arXiv preprint arXiv:1506.02211}}
  (\bibinfo{year}{2015}).
\newblock


\bibitem[Dosovitskiy et~al\mbox{.}(2020)]%
        {dosovitskiy2020image}
\bibfield{author}{\bibinfo{person}{Alexey Dosovitskiy}, \bibinfo{person}{Lucas
  Beyer}, \bibinfo{person}{Alexander Kolesnikov}, \bibinfo{person}{Dirk
  Weissenborn}, \bibinfo{person}{Xiaohua Zhai}, \bibinfo{person}{Thomas
  Unterthiner}, \bibinfo{person}{Mostafa Dehghani}, \bibinfo{person}{Matthias
  Minderer}, \bibinfo{person}{Georg Heigold}, \bibinfo{person}{Sylvain Gelly},
  {et~al\mbox{.}}} \bibinfo{year}{2020}\natexlab{}.
\newblock \showarticletitle{An image is worth 16x16 words: Transformers for
  image recognition at scale}.
\newblock \bibinfo{journal}{\emph{arXiv preprint arXiv:2010.11929}}
  (\bibinfo{year}{2020}).
\newblock


\bibitem[Fang et~al\mbox{.}(2021)]%
        {fang2021read}
\bibfield{author}{\bibinfo{person}{Shancheng Fang}, \bibinfo{person}{Hongtao
  Xie}, \bibinfo{person}{Yuxin Wang}, \bibinfo{person}{Zhendong Mao}, {and}
  \bibinfo{person}{Yongdong Zhang}.} \bibinfo{year}{2021}\natexlab{}.
\newblock \showarticletitle{Read like humans: Autonomous, bidirectional and
  iterative language modeling for scene text recognition}. In
  \bibinfo{booktitle}{\emph{Proceedings of the IEEE/CVF Conference on Computer
  Vision and Pattern Recognition}}. \bibinfo{pages}{7098--7107}.
\newblock


\bibitem[Flusser et~al\mbox{.}(2015)]%
        {flusser2015recognition}
\bibfield{author}{\bibinfo{person}{Jan Flusser}, \bibinfo{person}{Sajad
  Farokhi}, \bibinfo{person}{Cyril H{\"o}schl},
  \bibinfo{person}{Tom{\'a}{\v{s}} Suk}, \bibinfo{person}{Barbara Zitova},
  {and} \bibinfo{person}{Matteo Pedone}.} \bibinfo{year}{2015}\natexlab{}.
\newblock \showarticletitle{Recognition of images degraded by Gaussian blur}.
\newblock \bibinfo{journal}{\emph{IEEE transactions on Image Processing}}
  \bibinfo{volume}{25}, \bibinfo{number}{2} (\bibinfo{year}{2015}),
  \bibinfo{pages}{790--806}.
\newblock


\bibitem[Ge et~al\mbox{.}(2020)]%
        {ge2020look}
\bibfield{author}{\bibinfo{person}{Shiming Ge}, \bibinfo{person}{Kangkai
  Zhang}, \bibinfo{person}{Haolin Liu}, \bibinfo{person}{Yingying Hua},
  \bibinfo{person}{Shengwei Zhao}, \bibinfo{person}{Xin Jin}, {and}
  \bibinfo{person}{Hao Wen}.} \bibinfo{year}{2020}\natexlab{}.
\newblock \showarticletitle{Look one and more: Distilling hybrid order
  relational knowledge for cross-resolution image recognition}. In
  \bibinfo{booktitle}{\emph{Proceedings of the AAAI conference on artificial
  intelligence}}, Vol.~\bibinfo{volume}{34}. \bibinfo{pages}{10845--10852}.
\newblock


\bibitem[He et~al\mbox{.}(2016)]%
        {he2016deep}
\bibfield{author}{\bibinfo{person}{Kaiming He}, \bibinfo{person}{Xiangyu
  Zhang}, \bibinfo{person}{Shaoqing Ren}, {and} \bibinfo{person}{Jian Sun}.}
  \bibinfo{year}{2016}\natexlab{}.
\newblock \showarticletitle{Deep residual learning for image recognition}. In
  \bibinfo{booktitle}{\emph{Proceedings of the IEEE conference on computer
  vision and pattern recognition}}. \bibinfo{pages}{770--778}.
\newblock


\bibitem[Hinton et~al\mbox{.}(2015)]%
        {hinton2015distilling}
\bibfield{author}{\bibinfo{person}{Geoffrey Hinton}, \bibinfo{person}{Oriol
  Vinyals}, {and} \bibinfo{person}{Jeff Dean}.}
  \bibinfo{year}{2015}\natexlab{}.
\newblock \showarticletitle{Distilling the knowledge in a neural network}.
\newblock \bibinfo{journal}{\emph{arXiv preprint arXiv:1503.02531}}
  (\bibinfo{year}{2015}).
\newblock


\bibitem[Hochreiter and Schmidhuber(1997)]%
        {hochreiter1997long}
\bibfield{author}{\bibinfo{person}{Sepp Hochreiter} {and}
  \bibinfo{person}{J{\"u}rgen Schmidhuber}.} \bibinfo{year}{1997}\natexlab{}.
\newblock \showarticletitle{Long short-term memory}.
\newblock \bibinfo{journal}{\emph{Neural computation}} \bibinfo{volume}{9},
  \bibinfo{number}{8} (\bibinfo{year}{1997}), \bibinfo{pages}{1735--1780}.
\newblock


\bibitem[Huang et~al\mbox{.}(2018)]%
        {huang2018knowledge}
\bibfield{author}{\bibinfo{person}{Mingkun Huang}, \bibinfo{person}{Yongbin
  You}, \bibinfo{person}{Zhehuai Chen}, \bibinfo{person}{Yanmin Qian}, {and}
  \bibinfo{person}{Kai Yu}.} \bibinfo{year}{2018}\natexlab{}.
\newblock \showarticletitle{Knowledge Distillation for Sequence Model.}. In
  \bibinfo{booktitle}{\emph{Interspeech}}. \bibinfo{pages}{3703--3707}.
\newblock


\bibitem[Jaderberg et~al\mbox{.}(2014)]%
        {jaderberg2014speeding}
\bibfield{author}{\bibinfo{person}{Max Jaderberg}, \bibinfo{person}{Andrea
  Vedaldi}, {and} \bibinfo{person}{Andrew Zisserman}.}
  \bibinfo{year}{2014}\natexlab{}.
\newblock \showarticletitle{Speeding up convolutional neural networks with low
  rank expansions}.
\newblock \bibinfo{journal}{\emph{arXiv preprint arXiv:1405.3866}}
  (\bibinfo{year}{2014}).
\newblock


\bibitem[Jia et~al\mbox{.}(2021)]%
        {jia2021ifr}
\bibfield{author}{\bibinfo{person}{Zhiwei Jia}, \bibinfo{person}{Shugong Xu},
  \bibinfo{person}{Shiyi Mu}, \bibinfo{person}{Yue Tao}, \bibinfo{person}{Shan
  Cao}, {and} \bibinfo{person}{Zhiyong Chen}.} \bibinfo{year}{2021}\natexlab{}.
\newblock \showarticletitle{IFR: Iterative Fusion Based Recognizer for Low
  Quality Scene Text Recognition}. In \bibinfo{booktitle}{\emph{Pattern
  Recognition and Computer Vision: 4th Chinese Conference, PRCV 2021, Beijing,
  China, October 29--November 1, 2021, Proceedings, Part II 4}}. Springer,
  \bibinfo{pages}{180--191}.
\newblock


\bibitem[Karatzas et~al\mbox{.}(2015)]%
        {DimosthenisKaratzas2015ICDAR2C}
\bibfield{author}{\bibinfo{person}{Dimosthenis Karatzas},
  \bibinfo{person}{Lluis Gomez-Bigorda}, \bibinfo{person}{Anguelos Nicolaou},
  \bibinfo{person}{Suman~K. Ghosh}, \bibinfo{person}{Andrew~D. Bagdanov},
  \bibinfo{person}{Masakazu Iwamura}, \bibinfo{person}{Jiri Matas},
  \bibinfo{person}{Lukas Neumann}, \bibinfo{person}{Vijay Chandrasekhar},
  \bibinfo{person}{Shijian Lu}, \bibinfo{person}{Faisal Shafait},
  \bibinfo{person}{Seiichi Uchida}, {and} \bibinfo{person}{Ernest Valveny}.}
  \bibinfo{year}{2015}\natexlab{}.
\newblock \showarticletitle{ICDAR 2015 competition on Robust Reading}.
\newblock \bibinfo{journal}{\emph{International Conference on Document Analysis
  and Recognition}} (\bibinfo{year}{2015}).
\newblock


\bibitem[Karatzas et~al\mbox{.}(2013)]%
        {karatzas2013icdar}
\bibfield{author}{\bibinfo{person}{Dimosthenis Karatzas},
  \bibinfo{person}{Faisal Shafait}, \bibinfo{person}{Seiichi Uchida},
  \bibinfo{person}{Masakazu Iwamura}, \bibinfo{person}{Lluis~Gomez i Bigorda},
  \bibinfo{person}{Sergi~Robles Mestre}, \bibinfo{person}{Joan Mas},
  \bibinfo{person}{David~Fernandez Mota}, \bibinfo{person}{Jon~Almazan
  Almazan}, {and} \bibinfo{person}{Lluis~Pere De~Las~Heras}.}
  \bibinfo{year}{2013}\natexlab{}.
\newblock \showarticletitle{ICDAR 2013 robust reading competition}. In
  \bibinfo{booktitle}{\emph{2013 12th international conference on document
  analysis and recognition}}. IEEE, \bibinfo{pages}{1484--1493}.
\newblock


\bibitem[Khare et~al\mbox{.}(2019)]%
        {khare2019novel}
\bibfield{author}{\bibinfo{person}{Vijeta Khare},
  \bibinfo{person}{Palaiahnakote Shivakumara}, \bibinfo{person}{Chee~Seng
  Chan}, \bibinfo{person}{Tong Lu}, \bibinfo{person}{Liang~Kim Meng},
  \bibinfo{person}{Hon~Hock Woon}, {and} \bibinfo{person}{Michael
  Blumenstein}.} \bibinfo{year}{2019}\natexlab{}.
\newblock \showarticletitle{A novel character segmentation-reconstruction
  approach for license plate recognition}.
\newblock \bibinfo{journal}{\emph{Expert Systems with Applications}}
  \bibinfo{volume}{131} (\bibinfo{year}{2019}), \bibinfo{pages}{219--239}.
\newblock


\bibitem[Kim and Rush(2016)]%
        {kim2016sequence}
\bibfield{author}{\bibinfo{person}{Yoon Kim} {and} \bibinfo{person}{Alexander~M
  Rush}.} \bibinfo{year}{2016}\natexlab{}.
\newblock \showarticletitle{Sequence-level knowledge distillation}.
\newblock \bibinfo{journal}{\emph{arXiv preprint arXiv:1606.07947}}
  (\bibinfo{year}{2016}).
\newblock


\bibitem[Kingma and Ba(2014)]%
        {DiederikPKingma2014AdamAM}
\bibfield{author}{\bibinfo{person}{Diederik~P. Kingma} {and}
  \bibinfo{person}{Jimmy Ba}.} \bibinfo{year}{2014}\natexlab{}.
\newblock \showarticletitle{Adam: A Method for Stochastic Optimization}.
\newblock \bibinfo{journal}{\emph{arXiv: Learning}} (\bibinfo{year}{2014}).
\newblock


\bibitem[Ledig et~al\mbox{.}(2017)]%
        {ledig2017photo}
\bibfield{author}{\bibinfo{person}{Christian Ledig}, \bibinfo{person}{Lucas
  Theis}, \bibinfo{person}{Ferenc Husz{\'a}r}, \bibinfo{person}{Jose
  Caballero}, \bibinfo{person}{Andrew Cunningham}, \bibinfo{person}{Alejandro
  Acosta}, \bibinfo{person}{Andrew Aitken}, \bibinfo{person}{Alykhan Tejani},
  \bibinfo{person}{Johannes Totz}, \bibinfo{person}{Zehan Wang},
  {et~al\mbox{.}}} \bibinfo{year}{2017}\natexlab{}.
\newblock \showarticletitle{Photo-realistic single image super-resolution using
  a generative adversarial network}. In \bibinfo{booktitle}{\emph{Proceedings
  of the IEEE conference on computer vision and pattern recognition}}.
  \bibinfo{pages}{4681--4690}.
\newblock


\bibitem[Liem et~al\mbox{.}(2018)]%
        {liem2018fvi}
\bibfield{author}{\bibinfo{person}{Hoang~Danh Liem},
  \bibinfo{person}{Nguyen~Duc Minh}, \bibinfo{person}{Nguyen~Bao Trung},
  \bibinfo{person}{Hoang~Tien Duc}, \bibinfo{person}{Pham~Hoang Hiep},
  \bibinfo{person}{Doan~Viet Dung}, {and} \bibinfo{person}{Dang~Hoang Vu}.}
  \bibinfo{year}{2018}\natexlab{}.
\newblock \showarticletitle{Fvi: An end-to-end vietnamese identification card
  detection and recognition in images}. In \bibinfo{booktitle}{\emph{2018 5th
  NAFOSTED Conference on Information and Computer Science (NICS)}}. IEEE,
  \bibinfo{pages}{338--340}.
\newblock


\bibitem[Liu et~al\mbox{.}(2022)]%
        {liu2022perceiving}
\bibfield{author}{\bibinfo{person}{Hao Liu}, \bibinfo{person}{Bin Wang},
  \bibinfo{person}{Zhimin Bao}, \bibinfo{person}{Mobai Xue},
  \bibinfo{person}{Sheng Kang}, \bibinfo{person}{Deqiang Jiang},
  \bibinfo{person}{Yinsong Liu}, {and} \bibinfo{person}{Bo Ren}.}
  \bibinfo{year}{2022}\natexlab{}.
\newblock \showarticletitle{Perceiving stroke-semantic context: Hierarchical
  contrastive learning for robust scene text recognition}. In
  \bibinfo{booktitle}{\emph{Proceedings of the AAAI Conference on Artificial
  Intelligence}}, Vol.~\bibinfo{volume}{36}. \bibinfo{pages}{1702--1710}.
\newblock


\bibitem[Long et~al\mbox{.}(2015)]%
        {long2015learning}
\bibfield{author}{\bibinfo{person}{Mingsheng Long}, \bibinfo{person}{Yue Cao},
  \bibinfo{person}{Jianmin Wang}, {and} \bibinfo{person}{Michael Jordan}.}
  \bibinfo{year}{2015}\natexlab{}.
\newblock \showarticletitle{Learning transferable features with deep adaptation
  networks}. In \bibinfo{booktitle}{\emph{International conference on machine
  learning}}. PMLR, \bibinfo{pages}{97--105}.
\newblock


\bibitem[Ma et~al\mbox{.}(2023)]%
        {ma2023text}
\bibfield{author}{\bibinfo{person}{Jianqi Ma}, \bibinfo{person}{Shi Guo}, {and}
  \bibinfo{person}{Lei Zhang}.} \bibinfo{year}{2023}\natexlab{}.
\newblock \showarticletitle{Text prior guided scene text image
  super-resolution}.
\newblock \bibinfo{journal}{\emph{IEEE Transactions on Image Processing}}
  \bibinfo{volume}{32} (\bibinfo{year}{2023}), \bibinfo{pages}{1341--1353}.
\newblock


\bibitem[Ma et~al\mbox{.}(2022)]%
        {ma2022text}
\bibfield{author}{\bibinfo{person}{Jianqi Ma}, \bibinfo{person}{Zhetong Liang},
  {and} \bibinfo{person}{Lei Zhang}.} \bibinfo{year}{2022}\natexlab{}.
\newblock \showarticletitle{A Text Attention Network for Spatial Deformation
  Robust Scene Text Image Super-resolution}. In
  \bibinfo{booktitle}{\emph{Proceedings of the IEEE/CVF Conference on Computer
  Vision and Pattern Recognition}}. \bibinfo{pages}{5911--5920}.
\newblock


\bibitem[Mou et~al\mbox{.}(2020)]%
        {YongqiangMou2020PlugNetDA}
\bibfield{author}{\bibinfo{person}{Yongqiang Mou}, \bibinfo{person}{Lei Tan},
  \bibinfo{person}{Hui Yang}, \bibinfo{person}{Jingying Chen},
  \bibinfo{person}{Leyuan Liu}, \bibinfo{person}{Rui Yan}, {and}
  \bibinfo{person}{Yaohong Huang}.} \bibinfo{year}{2020}\natexlab{}.
\newblock \showarticletitle{PlugNet: Degradation Aware Scene Text Recognition
  Supervised by a Pluggable Super-Resolution Unit}.
\newblock \bibinfo{journal}{\emph{european conference on computer vision}}
  (\bibinfo{year}{2020}).
\newblock


\bibitem[Na et~al\mbox{.}(2022)]%
        {na2022multi}
\bibfield{author}{\bibinfo{person}{Byeonghu Na}, \bibinfo{person}{Yoonsik Kim},
  {and} \bibinfo{person}{Sungrae Park}.} \bibinfo{year}{2022}\natexlab{}.
\newblock \showarticletitle{Multi-modal text recognition networks: Interactive
  enhancements between visual and semantic features}. In
  \bibinfo{booktitle}{\emph{European Conference on Computer Vision}}. Springer,
  \bibinfo{pages}{446--463}.
\newblock


\bibitem[Nakaune et~al\mbox{.}(2021)]%
        {nakaune2021skeleton}
\bibfield{author}{\bibinfo{person}{Shimon Nakaune}, \bibinfo{person}{Satoshi
  Iizuka}, {and} \bibinfo{person}{Kazuhiro Fukui}.}
  \bibinfo{year}{2021}\natexlab{}.
\newblock \showarticletitle{Skeleton-aware text image super-resolution}. In
  \bibinfo{booktitle}{\emph{Proceedings of the 32nd British Machine Vision
  Conference, Online}}. \bibinfo{pages}{22--25}.
\newblock


\bibitem[Phan et~al\mbox{.}(2013)]%
        {TrungQuyPhan2013RecognizingTW}
\bibfield{author}{\bibinfo{person}{Trung~Quy Phan},
  \bibinfo{person}{Palaiahnakote Shivakumara}, \bibinfo{person}{Shangxuan
  Tian}, {and} \bibinfo{person}{Chew~Lim Tan}.}
  \bibinfo{year}{2013}\natexlab{}.
\newblock \showarticletitle{Recognizing Text with Perspective Distortion in
  Natural Scenes}.
\newblock \bibinfo{journal}{\emph{international conference on computer vision}}
  (\bibinfo{year}{2013}).
\newblock


\bibitem[Qi et~al\mbox{.}(2021)]%
        {qi2021multi}
\bibfield{author}{\bibinfo{person}{Lu Qi}, \bibinfo{person}{Jason Kuen},
  \bibinfo{person}{Jiuxiang Gu}, \bibinfo{person}{Zhe Lin}, \bibinfo{person}{Yi
  Wang}, \bibinfo{person}{Yukang Chen}, \bibinfo{person}{Yanwei Li}, {and}
  \bibinfo{person}{Jiaya Jia}.} \bibinfo{year}{2021}\natexlab{}.
\newblock \showarticletitle{Multi-scale aligned distillation for low-resolution
  detection}. In \bibinfo{booktitle}{\emph{Proceedings of the IEEE/CVF
  Conference on Computer Vision and Pattern Recognition}}.
  \bibinfo{pages}{14443--14453}.
\newblock


\bibitem[Qiao et~al\mbox{.}(2021)]%
        {qiao2021pimnet}
\bibfield{author}{\bibinfo{person}{Zhi Qiao}, \bibinfo{person}{Yu Zhou},
  \bibinfo{person}{Jin Wei}, \bibinfo{person}{Wei Wang}, \bibinfo{person}{Yuan
  Zhang}, \bibinfo{person}{Ning Jiang}, \bibinfo{person}{Hongbin Wang}, {and}
  \bibinfo{person}{Weiping Wang}.} \bibinfo{year}{2021}\natexlab{}.
\newblock \showarticletitle{Pimnet: a parallel, iterative and mimicking network
  for scene text recognition}. In \bibinfo{booktitle}{\emph{Proceedings of the
  29th ACM International Conference on Multimedia}}.
  \bibinfo{pages}{2046--2055}.
\newblock


\bibitem[Qin et~al\mbox{.}(2022)]%
        {qin2022scene}
\bibfield{author}{\bibinfo{person}{Rui Qin}, \bibinfo{person}{Bin Wang}, {and}
  \bibinfo{person}{Yu-Wing Tai}.} \bibinfo{year}{2022}\natexlab{}.
\newblock \showarticletitle{Scene Text Image Super-Resolution via Content
  Perceptual Loss and Criss-Cross Transformer Blocks}.
\newblock \bibinfo{journal}{\emph{arXiv preprint arXiv:2210.06924}}
  (\bibinfo{year}{2022}).
\newblock


\bibitem[Redmon et~al\mbox{.}(2016)]%
        {redmon2016you}
\bibfield{author}{\bibinfo{person}{Joseph Redmon}, \bibinfo{person}{Santosh
  Divvala}, \bibinfo{person}{Ross Girshick}, {and} \bibinfo{person}{Ali
  Farhadi}.} \bibinfo{year}{2016}\natexlab{}.
\newblock \showarticletitle{You only look once: Unified, real-time object
  detection}. In \bibinfo{booktitle}{\emph{Proceedings of the IEEE conference
  on computer vision and pattern recognition}}. \bibinfo{pages}{779--788}.
\newblock


\bibitem[Shi et~al\mbox{.}(2016)]%
        {shi2016end}
\bibfield{author}{\bibinfo{person}{Baoguang Shi}, \bibinfo{person}{Xiang Bai},
  {and} \bibinfo{person}{Cong Yao}.} \bibinfo{year}{2016}\natexlab{}.
\newblock \showarticletitle{An end-to-end trainable neural network for
  image-based sequence recognition and its application to scene text
  recognition}.
\newblock \bibinfo{journal}{\emph{IEEE transactions on pattern analysis and
  machine intelligence}} \bibinfo{volume}{39}, \bibinfo{number}{11}
  (\bibinfo{year}{2016}), \bibinfo{pages}{2298--2304}.
\newblock


\bibitem[Shin et~al\mbox{.}(2022)]%
        {shin2022teaching}
\bibfield{author}{\bibinfo{person}{Sungho Shin}, \bibinfo{person}{Joosoon Lee},
  \bibinfo{person}{Junseok Lee}, \bibinfo{person}{Yeonguk Yu}, {and}
  \bibinfo{person}{Kyoobin Lee}.} \bibinfo{year}{2022}\natexlab{}.
\newblock \showarticletitle{Teaching where to look: Attention similarity
  knowledge distillation for low resolution face recognition}. In
  \bibinfo{booktitle}{\emph{Computer Vision--ECCV 2022: 17th European
  Conference, Tel Aviv, Israel, October 23--27, 2022, Proceedings, Part XII}}.
  Springer, \bibinfo{pages}{631--647}.
\newblock


\bibitem[Shivakumara et~al\mbox{.}(2014)]%
        {PalaiahnakoteShivakumara2014ARA}
\bibfield{author}{\bibinfo{person}{Palaiahnakote Shivakumara},
  \bibinfo{person}{Anhar Risnumawan}, \bibinfo{person}{Chee~Seng Chan}, {and}
  \bibinfo{person}{Chew~Lim Tan}.} \bibinfo{year}{2014}\natexlab{}.
\newblock \showarticletitle{A robust arbitrary text detection system for
  natural scene images}.
\newblock \bibinfo{journal}{\emph{Expert Systems With Applications}}
  (\bibinfo{year}{2014}).
\newblock


\bibitem[Simonyan and Zisserman(2014)]%
        {simonyan2014very}
\bibfield{author}{\bibinfo{person}{Karen Simonyan} {and}
  \bibinfo{person}{Andrew Zisserman}.} \bibinfo{year}{2014}\natexlab{}.
\newblock \showarticletitle{Very deep convolutional networks for large-scale
  image recognition}.
\newblock \bibinfo{journal}{\emph{arXiv preprint arXiv:1409.1556}}
  (\bibinfo{year}{2014}).
\newblock


\bibitem[Tran and Ho-Phuoc(2019)]%
        {tran2019deep}
\bibfield{author}{\bibinfo{person}{Hanh~TM Tran} {and} \bibinfo{person}{Tien
  Ho-Phuoc}.} \bibinfo{year}{2019}\natexlab{}.
\newblock \showarticletitle{Deep laplacian pyramid network for text images
  super-resolution}. In \bibinfo{booktitle}{\emph{2019 IEEE-RIVF International
  Conference on Computing and Communication Technologies (RIVF)}}. IEEE,
  \bibinfo{pages}{1--6}.
\newblock


\bibitem[Tran et~al\mbox{.}(2017)]%
        {tran2017disentangled}
\bibfield{author}{\bibinfo{person}{Luan Tran}, \bibinfo{person}{Xi Yin}, {and}
  \bibinfo{person}{Xiaoming Liu}.} \bibinfo{year}{2017}\natexlab{}.
\newblock \showarticletitle{Disentangled representation learning gan for
  pose-invariant face recognition}. In \bibinfo{booktitle}{\emph{Proceedings of
  the IEEE conference on computer vision and pattern recognition}}.
  \bibinfo{pages}{1415--1424}.
\newblock


\bibitem[Vaswani et~al\mbox{.}(2017)]%
        {vaswani2017attention}
\bibfield{author}{\bibinfo{person}{Ashish Vaswani}, \bibinfo{person}{Noam
  Shazeer}, \bibinfo{person}{Niki Parmar}, \bibinfo{person}{Jakob Uszkoreit},
  \bibinfo{person}{Llion Jones}, \bibinfo{person}{Aidan~N Gomez},
  \bibinfo{person}{{\L}ukasz Kaiser}, {and} \bibinfo{person}{Illia
  Polosukhin}.} \bibinfo{year}{2017}\natexlab{}.
\newblock \showarticletitle{Attention is all you need}.
\newblock \bibinfo{journal}{\emph{Advances in neural information processing
  systems}}  \bibinfo{volume}{30} (\bibinfo{year}{2017}).
\newblock


\bibitem[Wang et~al\mbox{.}(2011)]%
        {KaiWang2011EndtoendST}
\bibfield{author}{\bibinfo{person}{Kai Wang}, \bibinfo{person}{Boris Babenko},
  {and} \bibinfo{person}{Serge Belongie}.} \bibinfo{year}{2011}\natexlab{}.
\newblock \showarticletitle{End-to-end scene text recognition}.
\newblock \bibinfo{journal}{\emph{international conference on computer vision}}
  (\bibinfo{year}{2011}).
\newblock


\bibitem[Wang et~al\mbox{.}(2020)]%
        {WenjiaWang2020SceneTI}
\bibfield{author}{\bibinfo{person}{Wenjia Wang}, \bibinfo{person}{Enze Xie},
  \bibinfo{person}{Xuebo Liu}, \bibinfo{person}{Wenhai Wang},
  \bibinfo{person}{Ding Liang}, \bibinfo{person}{Chunhua Shen}, {and}
  \bibinfo{person}{Xiang Bai}.} \bibinfo{year}{2020}\natexlab{}.
\newblock \showarticletitle{Scene Text Image Super-Resolution in the Wild}.
\newblock \bibinfo{journal}{\emph{european conference on computer vision}}
  (\bibinfo{year}{2020}).
\newblock


\bibitem[Wang et~al\mbox{.}(2019)]%
        {wang2019textsr}
\bibfield{author}{\bibinfo{person}{Wenjia Wang}, \bibinfo{person}{Enze Xie},
  \bibinfo{person}{Peize Sun}, \bibinfo{person}{Wenhai Wang},
  \bibinfo{person}{Lixun Tian}, \bibinfo{person}{Chunhua Shen}, {and}
  \bibinfo{person}{Ping Luo}.} \bibinfo{year}{2019}\natexlab{}.
\newblock \showarticletitle{Textsr: Content-aware text super-resolution guided
  by recognition}.
\newblock \bibinfo{journal}{\emph{arXiv preprint arXiv:1909.07113}}
  (\bibinfo{year}{2019}).
\newblock


\bibitem[Wang et~al\mbox{.}(2021)]%
        {wang2021two}
\bibfield{author}{\bibinfo{person}{Yuxin Wang}, \bibinfo{person}{Hongtao Xie},
  \bibinfo{person}{Shancheng Fang}, \bibinfo{person}{Jing Wang},
  \bibinfo{person}{Shenggao Zhu}, {and} \bibinfo{person}{Yongdong Zhang}.}
  \bibinfo{year}{2021}\natexlab{}.
\newblock \showarticletitle{From two to one: A new scene text recognizer with
  visual language modeling network}. In \bibinfo{booktitle}{\emph{Proceedings
  of the IEEE/CVF International Conference on Computer Vision}}.
  \bibinfo{pages}{14194--14203}.
\newblock


\bibitem[Xie et~al\mbox{.}(2022)]%
        {xie2022toward}
\bibfield{author}{\bibinfo{person}{Xudong Xie}, \bibinfo{person}{Ling Fu},
  \bibinfo{person}{Zhifei Zhang}, \bibinfo{person}{Zhaowen Wang}, {and}
  \bibinfo{person}{Xiang Bai}.} \bibinfo{year}{2022}\natexlab{}.
\newblock \showarticletitle{Toward Understanding WordArt: Corner-Guided
  Transformer for Scene Text Recognition}. In
  \bibinfo{booktitle}{\emph{Computer Vision--ECCV 2022: 17th European
  Conference, Tel Aviv, Israel, October 23--27, 2022, Proceedings, Part
  XXVIII}}. Springer, \bibinfo{pages}{303--321}.
\newblock


\bibitem[Xu et~al\mbox{.}(2017)]%
        {xu2017learning}
\bibfield{author}{\bibinfo{person}{Xiangyu Xu}, \bibinfo{person}{Deqing Sun},
  \bibinfo{person}{Jinshan Pan}, \bibinfo{person}{Yujin Zhang},
  \bibinfo{person}{Hanspeter Pfister}, {and} \bibinfo{person}{Ming-Hsuan
  Yang}.} \bibinfo{year}{2017}\natexlab{}.
\newblock \showarticletitle{Learning to super-resolve blurry face and text
  images}. In \bibinfo{booktitle}{\emph{Proceedings of the IEEE international
  conference on computer vision}}. \bibinfo{pages}{251--260}.
\newblock


\bibitem[Yang et~al\mbox{.}(2022)]%
        {yang2022masked}
\bibfield{author}{\bibinfo{person}{Zhendong Yang}, \bibinfo{person}{Zhe Li},
  \bibinfo{person}{Mingqi Shao}, \bibinfo{person}{Dachuan Shi},
  \bibinfo{person}{Zehuan Yuan}, {and} \bibinfo{person}{Chun Yuan}.}
  \bibinfo{year}{2022}\natexlab{}.
\newblock \showarticletitle{Masked generative distillation}. In
  \bibinfo{booktitle}{\emph{Computer Vision--ECCV 2022: 17th European
  Conference, Tel Aviv, Israel, October 23--27, 2022, Proceedings, Part XI}}.
  Springer, \bibinfo{pages}{53--69}.
\newblock


\bibitem[Yu et~al\mbox{.}(2020)]%
        {yu2020towards}
\bibfield{author}{\bibinfo{person}{Deli Yu}, \bibinfo{person}{Xuan Li},
  \bibinfo{person}{Chengquan Zhang}, \bibinfo{person}{Tao Liu},
  \bibinfo{person}{Junyu Han}, \bibinfo{person}{Jingtuo Liu}, {and}
  \bibinfo{person}{Errui Ding}.} \bibinfo{year}{2020}\natexlab{}.
\newblock \showarticletitle{Towards accurate scene text recognition with
  semantic reasoning networks}. In \bibinfo{booktitle}{\emph{Proceedings of the
  IEEE/CVF Conference on Computer Vision and Pattern Recognition}}.
  \bibinfo{pages}{12113--12122}.
\newblock


\bibitem[Zhang et~al\mbox{.}(2018)]%
        {zhang2018image}
\bibfield{author}{\bibinfo{person}{Yulun Zhang}, \bibinfo{person}{Kunpeng Li},
  \bibinfo{person}{Kai Li}, \bibinfo{person}{Lichen Wang},
  \bibinfo{person}{Bineng Zhong}, {and} \bibinfo{person}{Yun Fu}.}
  \bibinfo{year}{2018}\natexlab{}.
\newblock \showarticletitle{Image super-resolution using very deep residual
  channel attention networks}. In \bibinfo{booktitle}{\emph{Proceedings of the
  European conference on computer vision (ECCV)}}. \bibinfo{pages}{286--301}.
\newblock


\bibitem[Zhao et~al\mbox{.}(2021)]%
        {CairongZhao2021SceneTI}
\bibfield{author}{\bibinfo{person}{Cairong Zhao}, \bibinfo{person}{Shuyang
  Feng}, \bibinfo{person}{Brian~Nlong Zhao}, \bibinfo{person}{Zhijun Ding},
  \bibinfo{person}{Jun Wu}, \bibinfo{person}{Fumin Shen}, {and}
  \bibinfo{person}{Heng~Tao Shen}.} \bibinfo{year}{2021}\natexlab{}.
\newblock \showarticletitle{Scene Text Image Super-Resolution via Parallelly
  Contextual Attention Network}.
\newblock \bibinfo{journal}{\emph{acm multimedia}} (\bibinfo{year}{2021}).
\newblock


\bibitem[Zhao et~al\mbox{.}(2022)]%
        {MinyiZhao2022C3STISRST}
\bibfield{author}{\bibinfo{person}{Minyi Zhao}, \bibinfo{person}{Miao Wang},
  \bibinfo{person}{Fan Bai}, \bibinfo{person}{Bingjia Li}, \bibinfo{person}{Jie
  Wang}, {and} \bibinfo{person}{Shuigeng Zhou}.}
  \bibinfo{year}{2022}\natexlab{}.
\newblock \showarticletitle{C3-STISR: Scene Text Image Super-resolution with
  Triple Clues}.
\newblock \bibinfo{journal}{\emph{international joint conference on artificial
  intelligence}} (\bibinfo{year}{2022}).
\newblock


\bibitem[Zhu et~al\mbox{.}(2019)]%
        {zhu2019low}
\bibfield{author}{\bibinfo{person}{Mingjian Zhu}, \bibinfo{person}{Kai Han},
  \bibinfo{person}{Chao Zhang}, \bibinfo{person}{Jinlong Lin}, {and}
  \bibinfo{person}{Yunhe Wang}.} \bibinfo{year}{2019}\natexlab{}.
\newblock \showarticletitle{Low-resolution visual recognition via deep feature
  distillation}. In \bibinfo{booktitle}{\emph{ICASSP 2019-2019 IEEE
  International Conference on Acoustics, Speech and Signal Processing
  (ICASSP)}}. IEEE, \bibinfo{pages}{3762--3766}.
\newblock


\bibitem[Zhu et~al\mbox{.}(2023)]%
        {zhu2023improving}
\bibfield{author}{\bibinfo{person}{Shipeng Zhu}, \bibinfo{person}{Zuoyan Zhao},
  \bibinfo{person}{Pengfei Fang}, {and} \bibinfo{person}{Hui Xue}.}
  \bibinfo{year}{2023}\natexlab{}.
\newblock \showarticletitle{Improving Scene Text Image Super-Resolution via
  Dual Prior Modulation Network}.
\newblock \bibinfo{journal}{\emph{arXiv preprint arXiv:2302.10414}}
  (\bibinfo{year}{2023}).
\newblock


\end{thebibliography}

\end{document}